\documentclass[preprint,12pt]{elsarticle}

\usepackage{amssymb}
\usepackage{amsmath, amsfonts}
\usepackage{algorithmic}
\usepackage{algorithm}
\usepackage{array}
\usepackage{textcomp}
\usepackage{stfloats}
\usepackage{multirow}
\usepackage{bm}
\usepackage{url}
\usepackage{verbatim}
\usepackage{subcaption}
\usepackage[normalem]{ulem}  % erhan added for strikethrough text

\usepackage[dvipsnames]{xcolor}

\newcommand{\erhan}[1]{\textcolor{black}{#1}}

\journal{Neural Networks}

\begin{document}

\begin{frontmatter}



\title{%Energy Modulated Learning Progress Guided Interleaved Multi-Task Learning 
Interleaved Multitask Learning with Energy Modulated Learning Progress
\tnoteref{t1}}
%\tnotetext[t1]{This work is funded by the .....}


\author[1]{Hanne Say \corref{cor1}} %% Author name
\ead{hanne.say@ozu.edu.tr}

\author[2]{Suzan Ece Ada}
\ead{ece.ada@bogazici.edu.tr}

\author[2]{Emre Ugur}
\ead{emre.ugur@bogazici.edu.tr}

\author[3,4]{Minoru Asada}
\ead{asada@otri.osaka-u.ac.jp}

\author[1,3]{Erhan Oztop}
\ead{erhan.oztop@{otri.osaka-u.ac.jp, ozyegin.edu.tr}}

\cortext[cor1]{Corresponding author}
% \fntext[fn1]{This is the first author footnote.}
% \fntext[fn2]{Another author footnote, this is a very long footnote and
% it should be a really long footnote. But this footnote is not yet
% sufficiently long enough to make two lines of footnote text.}
% \fntext[fn3]{Yet another author footnote.}

\affiliation[1]{organization={Artificial Intelligence and Data Engineering Department, Ozyegin University},%Department and Organization 
            city={Istanbul},
            postcode={34794},
            country={Turkey}}

\affiliation[2]{organization={Department of Computer Engineering, Bogazici University},%Department and Organization
            city={Istanbul},
            postcode={34342}, 
            country={Turkey}}

\affiliation[3]{organization={SISREC, OTRI, University of Osaka},%Department and Organization
            city={Osaka},
            postcode={565-0871},
            country={Japan}}

\affiliation[4]{organization={International Professional University of Technology},%Department and Organization
            city={Osaka},
            postcode={530-0001},
            country={Japan}}

%% Abstract
\begin{abstract}
%Humans continuously acquire new skills and utilize existing knowledge while retaining the information that was previously learned. Analogously, in machine learning, `continual learning' seeks to assimilate new information while preserving past knowledge and leveraging from it where possible. 
As humans learn new skills and apply their existing knowledge while maintaining previously learned information, `continual learning' in machine learning aims to incorporate new data while retaining and utilizing past knowledge. However, existing machine learning methods often does not mimic  human learning where tasks are intermixed due to individual preferences and environmental conditions.  Humans typically switch between tasks instead of completely mastering one task before proceeding to the next. To explore how human-like task switching can enhance learning efficiency, we propose a multi task learning architecture that alternates tasks based on task-agnostic measures such as `learning progress' and `neural computational energy expenditure'. To evaluate the efficacy of our method, we run several systematic experiments by using a set of effect-prediction tasks executed by a simulated manipulator robot. The experiments show that our approach surpasses random interleaved and sequential task learning in terms of average learning accuracy. Moreover, by including energy expenditure in the task switching logic, our approach can still perform favorably while reducing neural energy expenditure.

\end{abstract}

%% Keywords

\end{frontmatter}



\section{Introduction}
Humans has the ability to learn continuously by integrating new information while exploiting and retaining previously acquired knowledge. This lifelong learning process is characterized by constant interleaving of tasks and goals. We switch tasks in response to environmental and cognitive constraints, adjusting our learning behavior in real time - for example, deciding to disengage from a task when it becomes less efficient or more mentally demanding.\cite{mittelstadt2021balancing}. This dynamic task switching not only enhances memory consolidation but also adapts us to dynamic environments by shifting attention toward differences between tasks and forcing us to be more mentally engaged \cite{firth2021systematic}.

Traditional machine learning approaches often rely on isolated task learning, which optimizes performance for a specific task without considering knowledge transfer from related tasks,  limiting cross-domain generalization and adaptability \cite{chen2018lifelong}. \emph{Multi-task learning}, on the other hand, aims to improve generalization by training models on multiple tasks simultaneously \cite{zhang2021survey}. In \emph{continual learning},  sequential task acquisition  is addressed by learning tasks one after another without forgetting the earlier ones and making use of the previous tasks where possible \cite{oztop2020lifelong,parisi2019continual},  which is called condensed learning in \cite{BAKER2023274}
%https://www.sciencedirect.com/science/article/pii/S0893608023000072 
as opposed to dispersed or interleaved learning. The order of the tasks however, affect the learning and transfer performance in these paradigms \cite{BAKER2023274, bell2022effect}. Instead of a static task scheduling, humans exhibit online dynamic task selection leading to an emergent interleaved task learning schedule.

The human brain has evolved to optimize energy efficiency, favoring strategies that maximize information transmission per unit energy \cite{yu2017energy}. With mechanisms such as synaptic pruning and neural plasticity \cite{kolb2011brain}, it optimizes its neural pathways, ensuring efficient cognitive function while reducing its metabolic costs \cite{bullmore2012economy, spear2013adolescent}.
Along with the brain’s drive toward efficient information processing, we also get inspiration from \emph{learning progress (LP)}, which has been proposed as an intrinsic motivation mechanism that guides exploration based on the rate of improvement in performance. Schmidhuber \cite{schmidhuber1991possibility, schmidhuber2006developmental} introduced LP as a measure of how quickly an agent reduces its prediction error, encouraging the agent to focus on experiences that yield the highest learning gains. This aligns well with biological principles, as the brain appears to allocate information encoding resources preferentially to stimuli that are neither too simple nor too complex, maximizing informational value relative to effort \cite{kidd2014goldilocks, kidd2010goldilocks}. 

In this paper, we propose a novel approach that integrates interleaved multi-task learning with energy-modulated learning progress (IMTL-EMLP) approach. By designing a system that not only has capacity to switch between tasks in an interleaved fashion but also adjusts its learning dynamics based on energy efficiency.
Our method is evaluated in a simulated robotic environment where the robot is tasked with learning the effects of its action in different environmental and task settings. 

The experimental results demonstrate that suitable task switching leads to an interleaved learning regime, which improves learning performance across different tasks.
Additionally, introduction of energy-modulation to the
learning progress based task selection results in reduced energy consumption without a significant drop in learning efficiency. Overall the contributions of our work is as follows:
\begin{itemize}
    \item First time multi-task learning using neural energy and learning progress for learning arbitration is realized leading to interleaved learning similar to that of humans  
    \item A novel multi-task learning architecture is introduced that uses shared attention layers to enable bi-directional skill transfer in interleaved learning 
    \item Realization of the proposed interleaved multi-task learning framework on effect prediction tasks in a simulated robotic environment.
    \item  Empirical demonstration that integrating energy-aware learning progress criteria reduces neural energy expenditure without degrading the agent’s learning performance
\end{itemize}
The remainder of the paper is organized as follows: Section~\ref{sec:lit} reviews related work on multi-task learning, interleaved learning, intrinsic motivation, and energy-efficient computation; Section~\ref{sec:prob} formulates the problem of dynamic task arbitration formally; Section~\ref{sec:method} introduces our proposed interleaved multi-task learning framework with energy-modulated learning progress; Section~\ref{sec:exp} presents the experimental setup and evaluates the performance of our approach across various metrics and baselines; and Section~\ref{sec:conc} concludes the paper with a summary of findings and directions for future research.

\section{Related Work}\label{sec:lit}

\subsection{Multi-task Learning}
Multi-task learning (MTL) \cite{caruana1997multitask} is a machine learning paradigm in which several tasks are learned jointly so that knowledge contained in one task can help improve the performance of the others. Unlike transfer learning—where the emphasis is on improving one target task, MTL treats all tasks as equally important and seeks to boost them collectively. MTL is also distinct from multi-label learning and multi-output regression because in MTL, the tasks typically do not share the same data records (i.e., each task has its own dataset). The key questions pertaining to MTL are (1) when to use MTL to share knowledge, (2) what to share (features, instances, or parameters), and (3) how to share, e.g., by feature learning, clustering, or parameter sharing. The most commonly studied question in the literature is ``how" \cite{zhang2021survey}, as we discuss next.

In feature learning, it is common to map the original features \cite{argyriou2008convex, maurer2013sparse, liu2017adversarial, shinohara2016adversarial} to a common space or choose a subset of the original task features \cite{lee2010adaptive, wang2015safe}. One of the notable architectures in the feature learning category is the Cross-Stitch Networks \cite{misra2016cross}. It dynamically combines feature maps from parallel task-specific networks, by a \emph{cross-stitch unit} that learns a linear combination of the activation maps from the previous layer, allowing the model to adaptively determine which features to share based on the input data. Cross-Stitch Networks has inspired our earlier skill transfer model \cite{say2023model} and shares conceptual elements with the proposed model of the current study. Yet, our model differs in several critical aspects. Cross-Stitch Networks are typically initialized with task-specific single-task networks, finetuned on the respective task, and then combined using cross-stitch units to perform joint training. In contrast, in our model, all tasks are trained from scratch in a unified architecture without any pretraining or fine-tuning. In addition, after introducing the cross-stitch units, in their approach, all the tasks are trained \emph{jointly}, using the same input data, where we only train one task per training iteration, which is chosen by our task arbitration mechanism, paralleling human learning. % aiming human-like interleaved learning.   
Other than the feature-based models, parameter-based models are also developed for MTL problems such as low-rank methods \cite{pong2010trace, han2016multi} suggesting that if tasks are related, the matrix of model parameters often has low rank, task clustering methods \cite{crammer2012learning, barzilai2015convex, zhou2015flexible} that hypothesize tasks form separate clusters of similarity, with each group sharing parameters. Other parameter-based approaches, such as task relation \cite{zhang2017learning, lee2016asymmetric} and parameter matrix decomposition \cite{gong2012robust, NIPS2010_00e26af6}, are likewise covered in the literature. 

Multi-task learning is also addressed in reinforcement learning (RL), where the RL agent seeks to improve performance across various decision-making tasks by leveraging shared experiences and representations. A significant challenge in Multi-task RL (MTRL) is negative interference, where learning in one task can degrade performance in others \cite{teh2017distral}. To address negative transfer, Hallak \emph{et~al.} (2015) introduced Contextual Markov Decision Process (CMDP), wherein a context defines the parameters of the transition and reward functions, with each task associated with a context vector \cite{hallak2015contextual}. Subsequently, Sodhani \emph{et~al.} \cite{sodhani2021multi} extended this formalism to Block Contextual Markov Decision Process (BC-MDP), incorporating the state space into the context-dependent framework. 
Similarly, Hendawy \emph{et~al.} \cite{hendawy2024multitask} promote diversity among representations by learning orthogonal representations through the Gram-Schmidt process. 
On the other hand, Teh \emph{et~al.} \cite{teh2017distral} uses a distilled policy to learn a shared representation that guides the task-specific policies, where the task-specific structure aims to mitigate updates that can lead to negative interference. These methods jointly demonstrate the significance of structuring shared representations and alleviating negative interference in MTRL. In our model, we do this by using an attention layer to facilitate positive skill transfer and suppress negative interference.

An open question in MTL, other than learning and skill sharing, is how to balance the contributions of individual tasks that may vary widely in complexity and data distribution. Recent research has addressed this through adaptive loss weighting and gradient balancing. Kendall \emph{et~al.} \cite{kendall2018multi} introduced an uncertainty-based framework that scales each task’s loss according to its inherent noise, thereby directing the learning process toward more reliable signals. Complementing this, Chen \emph{et~al.} \cite{chen2018gradnorm} proposed GradNorm, a method that normalizes the gradients of different tasks to ensure that no single task dominates the optimization process. 

In addition, Sener and Koltun \cite{sener2018multitask} reframed MTL as a multi-objective optimization problem, seeking Pareto-optimal solutions that effectively navigate trade-offs between conflicting task objectives. MTL approaches adopting mixture-of-expert architectures are also proposed, such as in \cite{ma2018modeling}, where task contributions are dynamically weighted and shared representations are exploited. However, despite the effectiveness of these approaches in balancing learning objectives and modeling task relations, they typically assume simultaneous or sequential task training schedules. In contrast, our work departs from this by introducing a dynamic task arbitration mechanism that selects a single task at each training step based on learning progress and energy consumption. This enables an interleaved training regime that more closely resembles human learning patterns, which we describe in the following section.

\subsection{Interleaved Learning in Humans}
Interleaved learning is a cognitive strategy in which learners alternate among diverse topics or problem types rather than concentrating on one subject exclusively. 
Empirical research in cognitive psychology provides compelling evidence for the benefits of interleaved practice. For example, studies have demonstrated that interleaving mathematics problems enhance learners’ ability to distinguish between problem types, leading to improved problem-solving skills \cite{rohrer2007shuffling}. Similarly, research by Kornell and Bjork  \cite{kornell2008learning} indicates that interleaved learning facilitates the formation of more flexible and integrated representations of concepts, enabling learners to better apply acquired knowledge in new contexts. Enhanced retention and transfer are also observed while learning motor skills \cite{czyz2024high, LIN20181}.

Neuroimaging studies further show that interleaving practice drives increased frontal–parietal activity and heightened motor cortex excitability along with the reduced retrieval time of information, thus leading to better long-term retention and efficient retrieval compared to the blocked practice \cite{lin2011brain}. The positive effects of interleaved practice are usually associated with a phenomenon called \emph{contextual interference (CI)} \cite{shea1979contextual}, and considered as one of the `desirable difficulties' for learning \cite{christina1991optimizing}. This view suggests that introducing a challenge during learning can lead to an improvement in long-term retention \cite{schmidt1992new}. In \cite{rohrer2015interleaved}, math problems are shuffled so that the problems belonging to the same kind are not consecutively solved by the students, requiring students to come up with a proper strategy while solving the problem, similar to  real-world situations. The experiments showed that interleaved practice produced higher scores compared to blocked practice, in a final test given on both day 1 and day 30 (delayed test), showcasing protection against forgetting. In addition, the study suggests that interleaved learning benefits do not lessen over time; quite the opposite, they may increase over time. Similarly,  \cite{samani2021interleaved} showed that undergraduate physics students who practiced interleaved problem sets demonstrated improved memory and problem-solving skills compared to those who used blocked practice.
These findings highlight the potential of interleaved learning as a powerful strategy for optimizing human learning in both educational and real-world settings.

In machine learning, \erhan{ however, interleaved learning received very little attention.}
Recently, Mayo \emph{et~al.} \cite{mayo2023multitask} investigated the interleaving on a multi-task learning problem and discussed that rather than designing mechanisms to prevent forgetting, such as external memory \cite{kamra2017deep} or regularization of weights \cite{li2017learning, kirkpatrick2017overcoming}, we should focus on designing learning systems and schedules that embrace the natural and resilient mechanisms of human learning. In this view, forgetting is not a failure but a feature that coexists with the capacity for efficient recovery i.e., \emph{relearning savings} \cite{hinton1986learning} and long-term knowledge retention. Even without mechanisms to prevent forgetting, standard neural networks show memory retention effects \cite{mayo2023multitask}, similar to humans, when tasks are interleaved.

\subsection{Intrinsic Motivation and Learning Progress}

\emph{Intrinsic motivation (IM)} refers to the internal drive of an agent to engage in activities for their inherent satisfaction, rather than for external rewards \cite{ryan2000intrinsic}. In the context of AI and robotics, intrinsic motivation enables agents to exhibit behaviors such as curiosity \cite{gottlieb2013information, 6652525}, novelty \cite{1011821, cogprints2511}, and surprise \cite{achiam2017surprise}, similar to the motivations observed in humans \cite{white1959motivation, berlyne1960conflict, csikszentmihalyi1990flow}. As an example, in \cite{TANNEBERG201967},  cognitive dissonance, the mismatch between mental expectation and observation is used an IM signal to enable efficient online adaptation for robotic manipulation. Another IM that is often used in robotics and computational approaches is the prediction progress, also known as \emph{learning progress (LP)} \cite{oudeyer2007intrinsic}.
It is usually calculated by comparing the predictor's error before and after it is updated, using the same sensorimotor context \cite{schmidhuber1991possibility}. There are numerous applications of LP in machine learning problems such as exploration guiding \cite{lopes2012exploration}, region selection \cite{bugur2019effect, Sener2023}, and curriculum learning \cite{kanitscheider2021multi,colas2019curious} in RL. In a recent work, Ada \emph{et~al.} proposed a form in LP, namely, episodic return progress, for bidirectional progressive neural networks to facilitate skill transfer among morphologically different agents \cite{ada2024}. Similarly, Colas \emph{et~al.} used LP in their multi-goal RL model (CURIOUS) as an IM signal to select which goal module to practice and replay, prioritizing those with the highest absolute LP, producing an adaptive curriculum for learning \cite{colas2019curious}. In the same vein, in our previous work \cite{say2023model}, we used LP as a signal to autonomously arbitrate task selection. Unlike prior works that use LP solely for guiding exploration, goal or region selection, this study applies LP at the task level and integrates it with energy consumption to enable efficient, interleaved multi-task learning.

\subsection{Energy Conservation in the Brain}
Energy management in the human brain is not only about powering neural activity—it also reflects a sophisticated system for conserving energy that influences our behavior \cite{vergara2019energy}. Despite the brain’s relatively small size, it consumes a disproportionate share of the body's energy, mainly to support neural signaling through the maintenance of ion gradients and synaptic transmission \cite{attwell2001energy}. Over time, the brain has evolved mechanisms such as synaptic pruning and circuit rewiring, where frequently used pathways are strengthened, and rarely used connections are eliminated. This dynamic reorganization minimizes unnecessary activity, leading to what is often described as neural efficiency \cite{lennie2003cost, raichle2006brains}.
In addition to the brain's physiological mechanisms for preserving energy, research in cognitive neuroscience and psychology has also shown that people systematically avoid tasks perceived as highly demanding, requiring more effort or energy, making the associated rewards seem less valuable, which is often explained by the term \emph{effort discounting} \cite{botvinick2009effort}. For example, Kool \emph{et~al.} \cite{kool2010decision} demonstrated that individuals tend to choose less cognitively demanding tasks when given the option, reflecting an inherent preference for minimizing mental effort. Similarly, Westbrook and Braver \cite{westbrook2015cognitive} provided evidence that cognitive effort carries a subjective cost, influencing decision-making processes. This concept is further elaborated in the opportunity cost model proposed by Kurzban \emph{et~al.} \cite{kurzban2013opportunity}, which suggests that the perceived cost of expending cognitive energy plays a central role in our choices, leading us to opt for tasks that require minimal resource expenditure. In this study,  we are inspired by the findings in neural and cognitive sciences and propose an energy-efficient task selection mechanism in addition to learning-based task arbitration.

\subsection{Effect Prediction and its Applications}
Effect prediction tasks in robotics involve endowing agents with the capability to anticipate the outcomes of their actions, similar to humans \cite{flanagan2006control}, forming an internal model of the environment that can be used for planning and control \cite{montesano2008learning}. This predictive capacity—often referred to as a forward model \cite{mehta2002forward} is fundamental to intelligent behavior. By simulating the effects of different actions, a robot can evaluate potential strategies \cite{mericcli2015push} before committing to a course of action, thereby enhancing safety, efficiency, and adaptability.

At the core of effect prediction is the idea that the robot learns to map its actions to subsequent states. Early approaches focused on building explicit physical models \cite{li2004iterative, tassa2012synthesis}; however, recent advances have shifted towards learning these models directly from high-dimensional sensory data using deep learning techniques. For example, Watter \emph{et~al.} \cite{watter2015embed} introduced a latent dynamics model that transforms raw sensory inputs, such as images, into a compact, low-dimensional latent space where the dynamics of the environment are more predictable. In this latent space, the model approximates the effect of actions as locally linear transformations, which can be leveraged for \erhan{short horizon} planning.
Another work by Agrawal \emph{et~al.} \cite{agrawal2016learning} focuses on self-supervised learning of intuitive physics. Their system trains a robot to predict the outcome of simple interactions, like poking objects, without explicit supervision.
Over time, the model learns to infer the underlying physical properties of objects, which is crucial for manipulating unfamiliar items.

Furthermore, effect prediction is central to model-based reinforcement learning (RL). In model-based RL, the agent uses its learned internal model to simulate future trajectories. This allows the agent to plan over longer time horizons and generate synthetic experiences, thereby reducing the need for extensive trial-and-error interactions with the real environment. PlaNet \cite{hafner2019learning} and Dreamer \cite{hafner2019dream} use state space models to learn latent dynamics from images. PlaNet employs a latent overshooting objective and performs planning in the learned latent space \cite{hafner2019learning}, whereas Dreamer uses an actor-critic approach, backpropagating value gradients through its latent world model \cite{hafner2019dream}. The “World Models” approach by Ha and Schmidhuber \cite{ha2018world} learns a compact representation of the world that supports internal simulation and policy training \cite{taniguchi2023world}. This internal simulation capability enables agents to achieve higher sample efficiency and improved performance, particularly in complex or resource-constrained settings.

In robotic applications, effect prediction tasks play a pivotal role in different problems, such as navigation, manipulation, and autonomous planning. In manipulation, \cite{tekden2024object} emphasizes the role of object- and relation-centric representations in improving push effect prediction, showing how modeling inter-object dynamics allows robots to infer and control the consequences of their actions in cluttered scenes. This low-level understanding of physical interactions feeds directly into goal-directed planning, as seen in \cite{ye2020object}. Ahmetoglu \emph{et~al.} \cite{ahmetoglu2025symbolic} demonstrate how effect prediction supports symbolic manipulation planning by learning object and relational predicates, providing a bridge between physical action and abstract reasoning \cite{Ugur2025}. While these works primarily focus on manipulation and planning, the principles extend to navigation as well; Aktas \emph{et~al.} \cite{aktas2024multi} show that predicting the effects of partial action executions enables more flexible, multi-step planning, which is essential for navigating dynamic or partially observable environments. However, these studies neither addressed multi-task learning nor used an intrinsic motivation signal to guide the action selection.

\section{Problem Statement}\label{sec:prob}
The general problem addressed is how a task arbitration and skill transfer mechanism can lead to improved overall learning and skill transfer performance in multitask learning settings. The framework sought to mimic human learning, where the learner has to decide which task to engage in and when to disengage, effectively allowing the emergence of arbitrary interleaved learning regimes.

We focus on supervised learning problems in robotics, and seek an online multi-task learning model equipped with a task arbitration mechanism that enables the robot to decide which task to engage in for data sampling and learning with the goal of minimizing average prediction error over the tasks. Formally, we are given $m$ tasks $ \left\{\mathcal{T}_i\right\}_{i=1}^m$
and an allocated interaction period $N_{\mathcal{T}_i}$ for each task. When task $\mathcal{T}_i$ is engaged, it is allowed to execute a total of $N_{\mathcal{T}_i}$ interactions with the environment. After this period, the task arbitration mechanism chooses a new task to be engaged. The selected task is used in execution, data collection, and learning. Denoting the selected task with $\mathcal{T}_*$, the execution experience for task $\mathcal{T}_*$  can be represented in the form of input, output pairs 
\[\mathcal{D}_{\mathcal{T}_*} = \{(x_j^{\mathcal{T}_*}, y_j^{\mathcal{T}_*})\}_{j=1}^{N_{\mathcal{T}_*}},\] 
where \(x_j^{\mathcal{T}_*}\) represents input while executing task \(\mathcal{T}_*\), and \(y_j^{\mathcal{T}_i}\) is the corresponding output at sampling step $j$. Note that by setting $N_{\mathcal{T}_i}$ values differently, one can encode the expected relative difficulty of the tasks, as in the case for tasks with high-dimensional versus low-dimensional input spaces.
Once the agent gathers a mini-batch of experience, $\mathcal{D}_{\mathcal{T}_*}$, it updates the task specific parameters, $\theta^{\mathcal{T}_*}$ as well as the shared task parameters, $\theta^s$. This online sampling and learning process continues to minimize the prediction error for each task. Denoting the prediction for task \(\mathcal{T}_i\) and loss associated with it by  $
f_{\mathcal{T}_i}\bigl(x^{\mathcal{T}_i};\theta^s,\theta^{\mathcal{T}_i}\bigr)$, \(\mathcal{L}_{\mathcal{T}_i}\) respectively, the objective for the multi-task learning agent can be stated as finding  \(\theta^s, \{\theta^{\mathcal{T}_i}\}_{i=1}^{m}\) to minimize the total loss, $\mathcal{TL}$ across tasks:  
\[
\mathcal{TL} = \min_{\theta^s,\{\theta^{\mathcal{T}_i}\}} \sum_{i=1}^{m} \mathcal{L}_{\mathcal{T}_i}\left( \left\{ \left(f_{\mathcal{T}_i}(x_j^{\mathcal{T}_i}; \theta^s, \theta^{\mathcal{T}_i}),\, y_j^{\mathcal{T}_i}\right) \right\}_{j=1}^{N_{\mathcal{T}_i}} \right).
\]

The problem statement projected to the experiments conducted in this paper can be given as follows:
The learning agent is required to learn a total of \(m\) distinct effect prediction tasks by establishing a mapping from the state-action space to the effect space. For each task \(\mathcal{T}_i\), the state is represented as  $x^{\mathcal{T}_i} \in \mathbb{R}^{d_s^{\mathcal{T}_i}}$ and the action as  $a^{\mathcal{T}_i} \in \mathbb{R}^{d_a^{\mathcal{T}_i}}$, where \(d_s^{\mathcal{T}_i}\) and \(d_a^{\mathcal{T}_i}\) denote the dimensions of the state and action spaces of task \(\mathcal{T}_i\), respectively. The corresponding effect, which is the outcome of executing action \(a^{\mathcal{T}_i}\) in state \(x^{\mathcal{T}_i}\), is denoted by \(e^{\mathcal{T}_i} \in \mathbb{R}^{d_e^{\mathcal{T}_i}}\). Thus, the experience collected and used in learning is $\mathcal{D}_{\mathcal{T}_i} = \{((x_j^{\mathcal{T}_i}, a_j^{\mathcal{T}_i}), e_j^{\mathcal{T}_i})\}_{j=1}^{N_{\mathcal{T}_i}}$. 
The goal is to dynamically arbitrate data sampling and learning among the tasks so that for each task $\mathcal{T}_i$ action-effect prediction functions, $f_{\mathcal{T}_i} : \mathbb{R}^{d_s^{\mathcal{T}_i}} \times \mathbb{R}^{d_a^{\mathcal{T}_i}} \rightarrow \mathbb{R}^{d_e^{\mathcal{T}_i}}$, are learned. After learning, it is expected that for any state-action pair \((x^{\mathcal{T}_i}, a^{\mathcal{T}_i})\) the predicted effect \(\Tilde{e}^{\mathcal{T}_i} = f_{\mathcal{T}_i}(x^{\mathcal{T}_i}, a^{\mathcal{T}_i})\) closely approximates the actual effect \(e^{\mathcal{T}_i}\). 

\section{Methodology}\label{sec:method}
In this section, we give the details of the learning architecture and task switching mechanisms developed and used in the experiments.

\subsection{Neural Network Architecture}
\begin{figure}[!h]
\centering
\includegraphics[width=\textwidth]{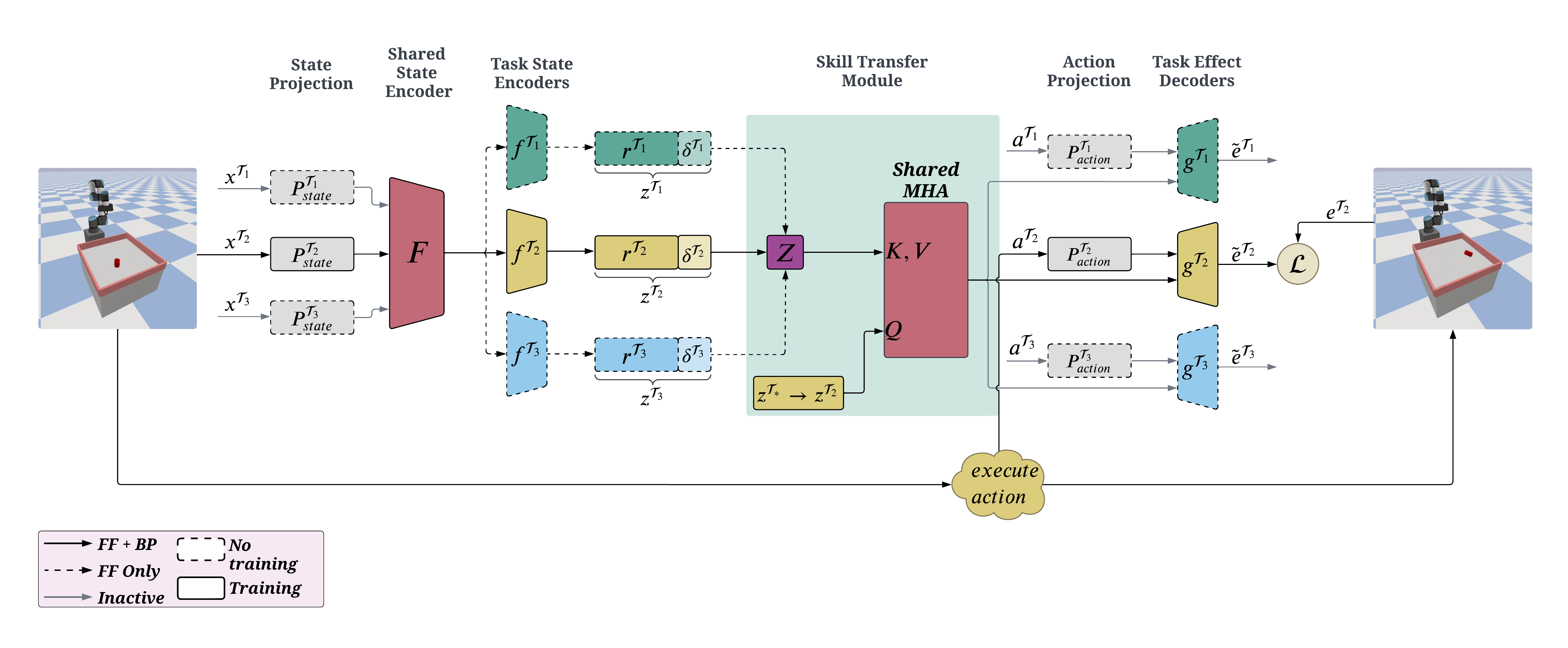}
\caption{Overview of the proposed multi-task effect prediction architecture and an example training iteration for $\mathcal{T}_*$ where $* = 2$ and total number of tasks $m = 3$. The model features shared and task-specific components, with only the selected task being trained at each step based on the arbitration mechanism.}
\label{fig:arch}
\end{figure}

Our model is based on an encoder-decoder network architecture designed to handle multiple-task learning %simultaneously 
while allowing for inter-task information sharing. The key components of the architecture are task-specific input (state) and action projection layers, task-specific encoders as well as a shared encoder and a shared attention module, and finally task-specific output (effect) decoders (see Figure~\ref{fig:arch}). We detail these components next. Symbols and notations used are defined in Table~\ref{tab:symbols} for clarity (see Table~\ref{tab:param} in the Appendix for the actual sizes).

\begin{table}[h]
\centering
\begin{tabular}{|c|p{11cm}|}
\hline
\textbf{Symbol} & \textbf{Description} \\
\hline
$m$ & Total number of tasks \\
\hline
$i$ & Task index, $\{\mathcal{T}_i\}_{i=1}^m$ \\
\hline
$\mathcal{T}_*$ & Selected task to be trained \\
\hline
$x^{\mathcal{T}_i}$ & State of task $\mathcal{T}_i$, $x^{\mathcal{T}_i} \in \mathbb{R}^{d_s^{\mathcal{T}_i}}$ \rule{0pt}{3ex} \\
\hline
$a^{\mathcal{T}_i}$ & Action of task $\mathcal{T}_i$, $a^{\mathcal{T}_i} \in \mathbb{R}^{d_a^{\mathcal{T}_i}}$ 
\rule{0pt}{3ex} \\
\hline
$e^{\mathcal{T}_i}$ & Effect of task $\mathcal{T}_i$, $e^{\mathcal{T}_i} \in \mathbb{R}^{d_e^{\mathcal{T}_i}}$ 
\rule{0pt}{3ex}\\
\hline
$P^{\mathcal{T}_i}_{\text{state}}$ & Task-specific state projection function, $P^{\mathcal{T}_i}_{\text{state}} : \mathbb{R}^{d_s^{\mathcal{T}_i}} \rightarrow \mathbb{R}^{d_s}$ \rule{0pt}{3ex}\\
\hline
$F$ & Shared state encoder, $F : \mathbb{R}^{d_s} \rightarrow \mathbb{R}^{d_h}$ \\
\hline
$h$ & Output of the shared encoder, $h \in \mathbb{R}^{d_h}$ \\
\hline
$r^{\mathcal{T}_i}$ & Task-specific latent representation, $r^{\mathcal{T}_i} \in \mathbb{R}^{d_r}$ 
\rule{0pt}{2.5ex}\\
\hline
$z^{\mathcal{T}_i}$ & Task-specific latent representation with flag bit, $z^{\mathcal{T}_i} \in \mathbb{R}^{d_r+1}$ 
\rule{0pt}{2.5ex}\\
\hline
$Z$ & Stacked matrix of all $\{z^{\mathcal{T}_i}\}_{i=1}^m$, $Z \in \mathbb{R}^{m \times (d_r+1)}$ 
\rule{0pt}{2.5ex}\\
\hline
$P^{\mathcal{T}_i}_{\text{action}}$ & Task-specific action projection function, $P^{\mathcal{T}_i}_{\text{action}} : \mathbb{R}^{d_a^{\mathcal{T}_i}} \rightarrow \mathbb{R}^{d_a}$ \rule{0pt}{3ex}\\
\hline
$f^{\mathcal{T}_i}$ & Task-specific encoder for task $\mathcal{T}_i$, $f^{\mathcal{T}_i} : \mathbb{R}^{d_h} \rightarrow \mathbb{R}^{d_r}$ 
\rule{0pt}{2.5ex}\\
\hline
$g^{\mathcal{T}_i}$ & Task-specific decoder for task $\mathcal{T}_i$, $g^{(t)} : \mathbb{R}^{d_r+d_a+1} \rightarrow \mathbb{R}^{d_e^{\mathcal{T}_i}}$ 
\rule{0pt}{3ex}\\
\hline
\end{tabular}
\caption{Overview of symbols and notations.}
\label{tab:symbols}
\end{table}

\begin{enumerate}
    \item{{\bf State Projection Layer.} For each task $\mathcal{T}_i$, the state input $x^{\mathcal{T}_i} \in \mathbb{R}^{d_s^{\mathcal{T}_i}}$ is first passed through a task-specific projection function $P_{state}^{\mathcal{T}_i} \colon \mathbb{R}^{d_s^{\mathcal{T}_i}} \to \mathbb{R}^{d_s}$ leading to the projected input vector given by:
        \begin{equation}
            \hat{x}^{\mathcal{T}_i} = P_{state}^{\mathcal{T}_i}(x^{\mathcal{T}_i}) ,
        \label{eq:state_proj}
        \end{equation}
    where $\hat{x}^{\mathcal{T}_i} \in \mathbb{R}^{d_s}$. This step ensures that inputs from different tasks have the same dimensionality for subsequent shared processing. The projection function may be chosen to be linear or nonlinear.  As the main objective of these layers is to have the same dimension at the output, a linear projection is used in the current work, as often done in the literature \cite{bengio2003neural, mikolov2013efficient}.}

    \item{{\bf Action Projection Layer.} Since the action vector $a^{\mathcal{T}_i} \in \mathbb{R}^{d_a^{\mathcal{T}_i}}$ for each task $t$ can vary in dimensionality, similar to the state projection layers, we project them to a fixed-dimensional space to maintain consistency across tasks. Each action vector is passed through a task-specific projection function $P_{action}^{\mathcal{T}_i} \colon \mathbb{R}^{d_a^{\mathcal{T}_i}} \to \mathbb{R}^{d_a}$:
        \begin{equation}
            \hat{a}^{\mathcal{T}_i} = P_{action}^{\mathcal{T}_i}(a^{\mathcal{T}_i}),
            \label{eq:act-proj}
        \end{equation}
    where $\hat{a}^{\mathcal{T}_i}\in\mathbb{R}^{d_a}$ is the projected action vector for task $\mathcal{T}_i$. This projection ensures that actions from different tasks are compatible for further processing. Similar to the \eqref{eq:state_proj}, we choose to have a linear projection layer for the action vectors.}

    \item{{\bf Shared State Encoder.} We incorporate a shared encoder in our neural network to encourage the extraction of general features that may be beneficial across multiple tasks, promoting knowledge sharing while reducing redundancy. Hence the projected input $\hat{x}^{\mathcal{T}_i}$ is fed  into the a shared encoder $F \colon \mathbb{R}^{d_s} \to \mathbb{R}^{d_h}$, to extract a $d_h$-dimensional  shared feature vector, $h\in\mathbb{R}^{d_h}$ with
    
        \begin{equation}
            h = F(\hat{x}^{\mathcal{T}_i}),
        \end{equation}
    }
    
    \item{{\bf Task State Encoders.} Following the shared encoder, we further integrate task-specific encoders, aiming to capture state knowledge unique to each task that the shared encoder cannot extract. We hypothesize that each task benefits from specialized features, enhancing its performance during training. Hence, $h$ is forwarded through all task-specific encoders in order to generate task-specific latent representations:
        \begin{equation}
           r^{\mathcal{T}_i} = f^{\mathcal{T}_i}(h), \quad \{\mathcal{T}_i\}_{i=1}^m
           \label{eq:encoder}
        \end{equation}
    Note that when training task \( \mathcal{T}_* \), only the parameters of \( f^{\mathcal{T}_*} \) are updated; all other \( f^{\mathcal{T}_i} \) modules remain frozen, with no gradient propagation.}
    
    \item{{\bf Shared Attention Module.}\label{sec: mha}} To facilitate inter-task communication and let the network focus on the current training task, we utilize a multi-head attention (MHA) mechanism \cite{vaswani2017attention}.
    Instead of having one attention function, where $Q, K, V \in \mathbb{R}^{d_{h}}$, MHA layer computes attention across $H$ parallel heads that each head projects $Q, K, V$ into a lower-dimensional space and performs scaled dot-product attention. Let $d_k$ be the dimensionality per head, and let $W_i^Q, W_i^K, W_i^V \in \mathbb{R}^{d_{h}\times d_k}$ be the projection matrices for head $i \in \{1,\dots,H\}$. Then, $Q, K, V$ are projected as follows:
    \begin{align*}
        Q_i = Q\,W_i^Q, 
        \quad
        K_i = K\,W_i^K, 
        \quad
        V_i = V\,W_i^V, \\
        \mathrm{head}_i = \mathrm{softmax}\!\Bigl(\frac{Q_i K_i^\top}{\sqrt{d_k}}\Bigr)\,V_i \in \mathbb{R}^{d_k}.
    \end{align*}
    
    The outputs from all heads are concatenated and passed through an output projection matrix to form the final attention output, denoted here as $A$, as follows:
    \[
    A = \left[\mathrm{head}_1 \; : \; \mathrm{head}_2 \; : \; \dots \; : \; \mathrm{head}_H\right] W^O \in d_{h},
    \]
    where $W^O \in \mathbb{R}^{(H \cdot d_k) \times d_{h}}$ is the output projection matrix and $[\cdot \; : \; \cdot]$ denotes vector concatenation. In \cite{vaswani2017attention}, all  $Q, K, V$ values have the same source in the \emph{self-attention} layers. In our work, we constrain the attention mechanism such that the current task is focused, or in other words `queried'. Concretely, when training task $\mathcal{T}_*$, we perform the following:

    \begin{itemize}
        \item Beforehand the attention function, we concatenate an additional flag bit to each task-specific latent representation, $r^{\mathcal{T}_i}$. Specifically, for each task $\mathcal{T}_i$, we concatenate a \emph{training flag} $\delta^{\mathcal{T}_i}$ indicating whether task $\mathcal{T}_i$ is currently being trained $(\delta^{\mathcal{T}_i}=1)$ or not $(\delta^{\mathcal{T}_i}=0)$. 
        The concatenated vector for each task is:
        \begin{equation}
            z^{\mathcal{T}_i}=[r^{\mathcal{T}_i}\; : \;\delta^{\mathcal{T}_i}] \in \mathbb{R}^{d_r+1} 
            \label{eq:repr}
        \end{equation}
        From now on, we call $z^{\mathcal{T}_i}$ the final representation of task $\mathcal{T}_i$.
        
        \item After gathering $z^{\mathcal{T}_i}$ for each task, they are stacked to form a matrix $Z$:
        \begin{equation}
            Z = \begin{bmatrix} z^{\mathcal{T}_1} \\ 
            z^{\mathcal{T}_2} \\ 
            \vdots \\ 
            z^{\mathcal{T}_m} 
            \end{bmatrix} \in \mathbb{R}^{m\times(d_r + 1)}.
            \label{eq:matrix}
        \end{equation}

        \item Once we obtain the matrix $Z$; $Q, K$ and $V$ values are gathered as below:
        \paragraph{Query ($Q$)} We choose the final representation of the current training task as the query since we want the attention mechanism to be driven by the perspective of the currently active task:
        \[
            Q = z^{\mathcal{T}_*} \in \mathbb{R}^{(d_r + 1)}.
        \]
        \paragraph{Key ($K$) and Value ($V$)} We use $Z$ so that the active training task can attend to representations from \emph{all} tasks (including itself):
        \[
            K = V = Z \in \mathbb{R}^{m\times (d_r + 1)}.
        \]
    \end{itemize}

     Using these $Q, K, V$ values, we feed them into our shared MHA module and get the attention output $A$ for the task $\mathcal{T}_*$:
    \begin{equation}
       A^{\mathcal{T}_*}  = SharedMHA(Q, K, V),
    \end{equation}
    where $A^{\mathcal{T}_*} \in \mathbb{R}^{(d_r + 1)}$.
    Because the query is restricted to the representation of the active task, $z^{\mathcal{T}_*}$, the network focuses on how the other tasks’ representations (contained in $Z$) can inform task $\mathcal{T}_*$. Moreover, the flags within each row of $Z$ indicate whether a given input originates from the actively trained task ($\delta^{\mathcal{T}_{i=*}}=1$) or from another task ($\delta^{\mathcal{T}_{i\neq*}}=0$), further guiding the attention toward balancing shared and task-specific information.

    \item{{\bf Task Effect Decoders.}} Same as the task-specific sub-encoder $f^{\mathcal{T}_i}$, for each task $\mathcal{T}_i$; we utilize a decoder module $g^{\mathcal{T}_i}$ designed for predicting the resulting effect (output) $e^{\mathcal{T}_i}$, after the executed action $a^{\mathcal{T}_i}$ while the environment is in the state $x^{\mathcal{T}_i}$. 
    For task $\mathcal{T}_*$, the output from the shared MHA mechanism $A^{\mathcal{T}_*}$ along with the projected action vector $\hat{a}^{\mathcal{T}_*}$ from \eqref{eq:act-proj} is given to the task-specific decoder of the selected task only, contrary to the \eqref{eq:encoder}, as follows:
    \begin{equation}
        \Tilde{e}^{\mathcal{T}_*} = g^{\mathcal{T}_*}([A^{\mathcal{T}_*}: \hat{a}^{\mathcal{T}_*}]).
    \end{equation}
    where $\Tilde{e}^{\mathcal{T}_*} \in \mathbb{R}^{d_e^{\mathcal{T}_*}}$. By integrating contextual information from the shared attention mechanism with the projected action vector, the decoder maps the state information to the corresponding effect.
\end{enumerate}

\subsection{Task Arbitration}
To investigate the effects of allowing interleaved learning in multi-task settings, we propose a task selection strategy based on the assessment of each task’s learning progress or its discounted version by the neural cost of learning. Unlike traditional multi-task learning approaches that train tasks in a fixed schedule or isolate them entirely, our method continuously evaluates the tasks’ performance trends and selects which task to train next based on these evaluations. By doing so, we aim to emulate the human tendency to switch among tasks. An overview of the proposed interleaved task engagement mechanism is presented in Algorithm \ref{alg:alg1}, where a non-greedy selection strategy is adopted by giving other tasks a chance to be selected, with an exploration rate $\epsilon$, thereby reducing over-reliance on the highest-scoring task.

    \begin{algorithm}[t]
    \caption{Interleaved Multi-Task Learning}\label{alg:alg1}
    \begin{algorithmic}
    \REQUIRE Number of tasks $m$, number of epochs $R$, exploration rate $\epsilon=0.1$
    \FOR{epoch $=1$ to $R$}
        \STATE Compute score $s^{\mathcal{T}_i}$ for tasks $\{\mathcal{T}_i\}_{i=1}^m$
        \STATE $i' \gets \underset{i}{\arg\max} (\{s^{\mathcal{T}_i}\}_{i=1}^m)$
        \STATE Generate a random number $r$ uniformly from $[0, 1]$
        \IF{$r < \epsilon$}
            \STATE $\mathcal{T}_* \gets$ Random task from $\{\mathcal{T}_1, \mathcal{T}_2, \ldots, \mathcal{T}_m\} \setminus \{\mathcal{T}_{i'}\}$
        \ELSE
            \STATE $\mathcal{T}_* \gets \mathcal{T}_{i'}$
        \ENDIF
        \STATE Perform training on the selected task $\mathcal{T}_*$
    \ENDFOR
    \end{algorithmic}
    \label{alg1}
    \end{algorithm}
    
\begin{enumerate}
    \item {\bf{Learning Progress (LP) Based Task Selection} } LP of a task is computed based  on the recent history of the error signal pertaining to it:  the error in the last $L$ ($L=5$ in the current study) steps linearly regressed against time and negative of its slope is taken as the LP value. To be concrete, let $E^{\mathcal{T}_i}_{t}$ be the error at time step $t$ for  task $\mathcal{T}_i$. 
    Consider the set of errors $\{E^{\mathcal{T}_i}_{t-L+1}, E^{\mathcal{T}_i}_{t-L+2}, \ldots, E^{\mathcal{T}_i}_{t}\}$. Then, a linear model is fit to these points to estimate the slope $\beta^{\mathcal{T}_i}$. If $\beta^{\mathcal{T}_i} < 0$, the error is decreasing, and the task is making positive learning progress. If $\beta^{\mathcal{T}_i} \geq 0$, it suggests that the error has plateaued or is increasing, implying no recent improvement. Under the LP-based interleaving scheme, the proposed task switching mechanism computes the learning progress $LP^{\mathcal{T}_i}$ for each task $\{\mathcal{T}_i\}_{i=1}^m$. Then it selects the task with the highest LP to be trained next. In other words, tasks that are currently showing rapid improvement receive more training time, while tasks that have stalled or regressed receive less immediate attention. 
    Formally, if $\beta^{\mathcal{T}_i}$ is the slope for task $\mathcal{T}_i$, then $\mathcal{T}_*$ is chosen in two steps:
    \begin{enumerate}
        \item 
        \begin{equation*}
            {LP^{\mathcal{T}_i}} = \begin{cases}
            \lvert \beta^{\mathcal{T}_i} \rvert,&{\text{if }}\ \beta^{\mathcal{T}_i} < 0 \\ 
            {0,}&{\text{otherwise.}} 
            \end{cases}
        \end{equation*}
        \item Select the task $\mathcal{T}_* = \underset{i}{\arg\max} (LP^{\mathcal{T}_i})$.
    \end{enumerate}
    This ensures that at any given moment, the model focuses on tasks where additional practice can yield meaningful gains, rather than spending time on tasks that have stagnated.
    
    \item {\bf{Energy Consumption Modulated Learning Progress (EMLP) Based Task Selection.}}
    While learning progress alone can guide the model toward tasks showing improvement, it does not consider the computational or energetic cost associated with training each task. In scenarios where energy efficiency is a concern, we incorporate energy consumption into the interleaving strategy. We define {\emph{energy consumption}} $EC^{\mathcal{T}_i}$ for task $\mathcal{T}_i$ as the total output of all neurons belonging to it $\mathcal{A}^{\mathcal{T}_i}$ , cumulated over the last $L$ training steps:
    \[
    EC^{\mathcal{T}_i} = \sum_{j=(t-L)}^t{\mathcal{A}}^{\mathcal{T}_i}_j.
    \]
    This could be measured in terms of computational operations, memory usage, or any proxy for energy expenditure. Lower $EC^{\mathcal{T}_i}$ values mean the task has recently been trained efficiently, while higher values indicate that the task has been relatively costly in terms of energy. However, using energy consumption only as the interleaving guidance, the network will only choose the tasks that produce low activation energy without taking the learning performance of the tasks into consideration, which can lead to worsened overall multi-task learning performance (see Figure~\ref{fig:k-loss}).
    To consider the cost of learning, LP is discounted by a positive constant $k$ that controls the sensitivity of the combined criteria to energy consumption:
    \begin{equation}
    \label{eq:energydiscount}
        s^{\mathcal{T}_i} = e^{(k \cdot \tilde{LP}^{\mathcal{T}_i})} / \tilde{EC}^{\mathcal{T}_i}.
    \end{equation}
    Note that, to combine LP and EC properly, they are first scaled between $[0, 1]$, before the combined score is computed, simply by using a min-max scaler.
    The $s^{\mathcal{T}_i}$ can be interpreted as follows:
    When $k$ approaches to zero, $e^{(k \cdot \tilde{LP}^{\mathcal{T}_i})} \approx 1$, thus the score is primarily influenced by $\tilde{EC}^{\mathcal{T}_i}$, even if the task has strong LP. This discourages spending too many `energy resources' on a single, costly task, even if it has high learning gains. On the other hand, when $k$ is high, $e^{(k \cdot \tilde{LP}^{\mathcal{T}_i})}$ becomes dominant, thus EC becomes negligible, approximating the previous LP-only selection. 
    Under the EC-modulated LP-based interleaving strategy, we compute combined scores for each task $\mathcal{T}_i$ as given above. The next task to train is the one with the highest combined score:
    \begin{equation}
        \mathcal{T}_* = \underset{i}{\arg\max} (s^{\mathcal{T}_i}).
    \end{equation}
    This ensures a balance between quick gains in task performance and maintaining overall energy efficiency. Tasks with high learning progress but also low recent energy costs are favored, while tasks that have become too energy-intensive or are not showing improvement receive less immediate focus. By integrating these interleaving strategies, we encourage a more dynamic, bio-inspired learning schedule. 
\end{enumerate}

\section{Experiments and Results}\label{sec:exp}
To evaluate the effectiveness of our proposed Interleaved Multitask Learning with Energy Modulated Learning Progress (IMTL-EMLP), a series of experiments are conducted with a simulated robot tasked with learning the effect of its actions in different task settings. In the experiments, we focus on learning efficiency and final task accuracy as performance measures and asses the knowledge transfer among tasks.

The proposed IMTL framework introduces Computational Energy modulated LP (EMLP) for dynamic task selection, which includes LP-based selection as a special case. We begin by evaluating the LP-based approach (IMTL-LP), followed by its energy-aware extension (IMTL-EMLP), which incorporates energy consumption into the task arbitration process.

To isolate the effects of task arbitration strategies, we compare IMTL-LP against several baselines: single-task learning (SINGLE), a multi-task learning model with random task selection (IMTL-RAND), and blocked learning setup where tasks are trained sequentially in fixed permutations (BLOCK). Additionally, we examine IMTL-EMLP, which extends IMTL-LP with energy sensitivity, analyzing how varying the energy coefficient $k$ affects the balance between prediction accuracy and computational cost.

All baselines use a consistent model architecture\footnote{The actual layer sizes for both single-task and multi-task learning setups are provided in the Appendix \ref{tab:param}. } to ensure that performance differences stem solely from task arbitration strategies. In addition to learning efficiency analysis, we assess the robustness to model size by comparing performance across low, medium, and high-resourced networks. Finally, two ablation studies dissect the individual and combined contributions of model components and explore object-wise skill transfer between tasks.

\subsection{Simulation Environment}
In this study, UR10 robotic arm equipped with a Franka Panda end effector is simulated via the PyBullet engine \cite{coumans2020}. We use a tabletop environment to evaluate multi-task learning through effect prediction tasks. 
A set of basic objects are presented on the table for the robot to interact with during the tasks. Namely,
a sphere (\( r = 3 \, \text{cm} \)), a cube (\( l = 6 \, \text{cm} \)), a square prism (\( w = 6 \, \text{cm}, \, h = 12 \, \text{cm} \)), and a cylinder (\( r = 3 \, \text{cm}, \, h = 18 \, \text{cm} \)) are used. The latter two objects can be presented in either vertical or horizontal orientation, thus making it six total object configurations for interaction. In each interaction, the object is randomly placed on the table.

\subsection{Effect Prediction Tasks}
For testing the proposed model, we have defined three tasks for the robotic agent as to predict the effect of its actions in three different settings as described next.
\\ \textbf{Push task.} A single object is randomly placed on the table. The robot applies a ``pushing" action with its end effector to the object’s center of mass (CoM), with an angle $\theta$ chosen between $[0, 180]$ degrees. The state of the object is taken as its pose defined by the Cartesian coordinates (x, y, z) and orientation (with Euler angles $\phi_x, \phi_y, \phi_z$). The orientation angles are represented with sine cosine pairs to avoid discontinuous input for the neural networks that process the object state. Thus, the object is represented as a 9-dimensional vector:
\begin{align*}
    \mathbf{s} = [x, y, z, \sin(\phi_x), \cos(\phi_x), 
    \sin(\phi_y), \cos(\phi_y), \sin(\phi_z), \cos(\phi_z)].
\end{align*}

The action is encoded using the sine and cosine of the push angle, combined with a one-hot vector indicating the object’s id:
\[
\mathbf{a} = [\sin(\theta), \cos(\theta), \text{onehot}(o)].
\]

The effect is defined as the change between the object's pre-action and post-action states, captured using the simulation environment: 
\begin{align*}
\mathbf{e} = [\Delta(x), \Delta(y), \Delta(z), \Delta(\sin(\phi_x)), \Delta(\cos(\phi_x)), \\
    \Delta(\sin(\phi_y)), \Delta(\cos(\phi_y)), \Delta(\sin(\phi_z)), \Delta(\cos(\phi_z))].
\end{align*}

\begin{figure*}[t]
    \captionsetup[subfloat]{labelfont=scriptsize,textfont=scriptsize} 
    \centering
    \subfloat[Push task.]{\includegraphics[scale=0.12]{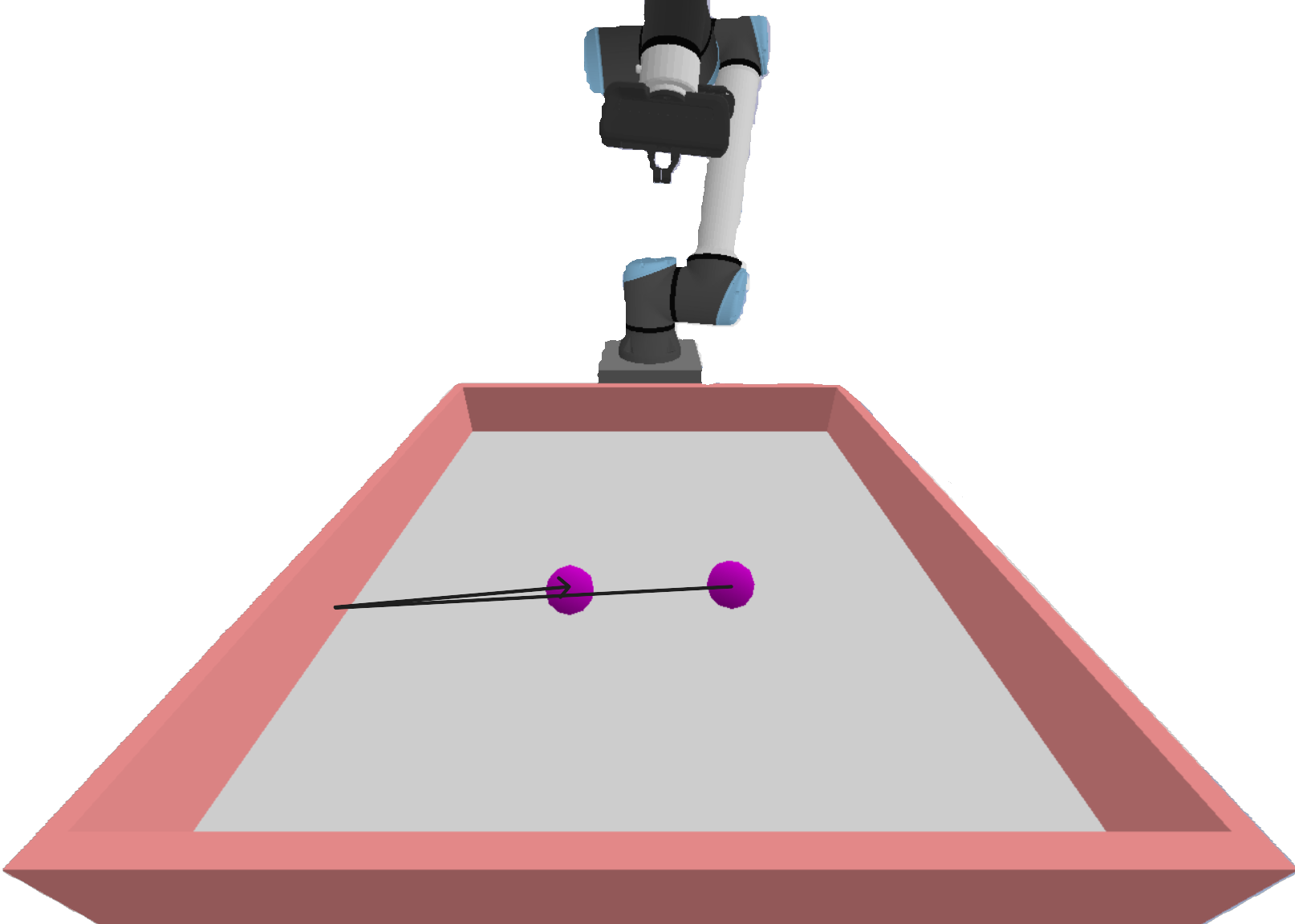}%
    \label{fig:push}}
    \hfill
    \subfloat[Hit task.]{\includegraphics[scale=0.12]{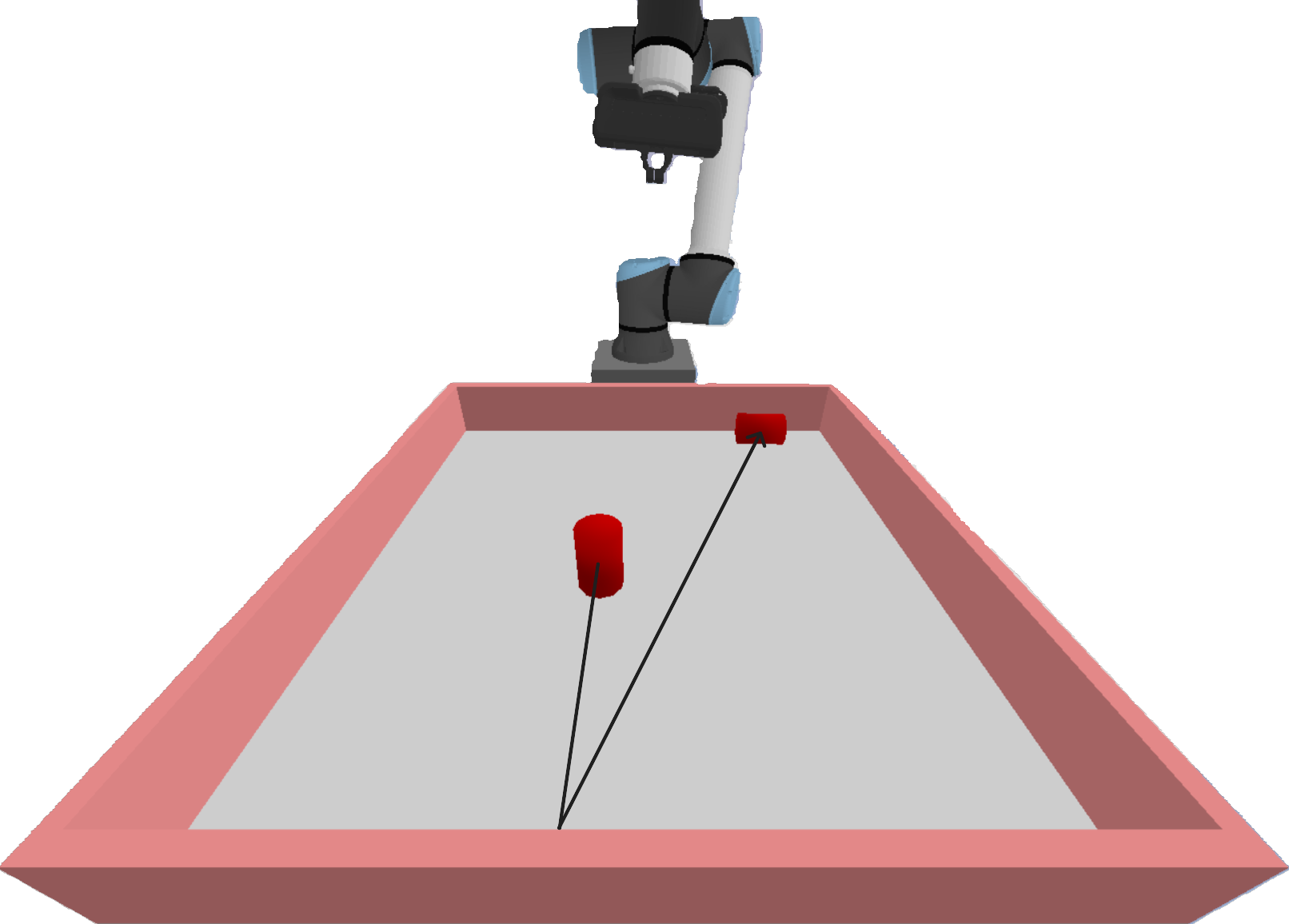}%
    \label{fig:hit}}
    \hfill
    \subfloat[Stack task.]{\includegraphics[scale=0.12]{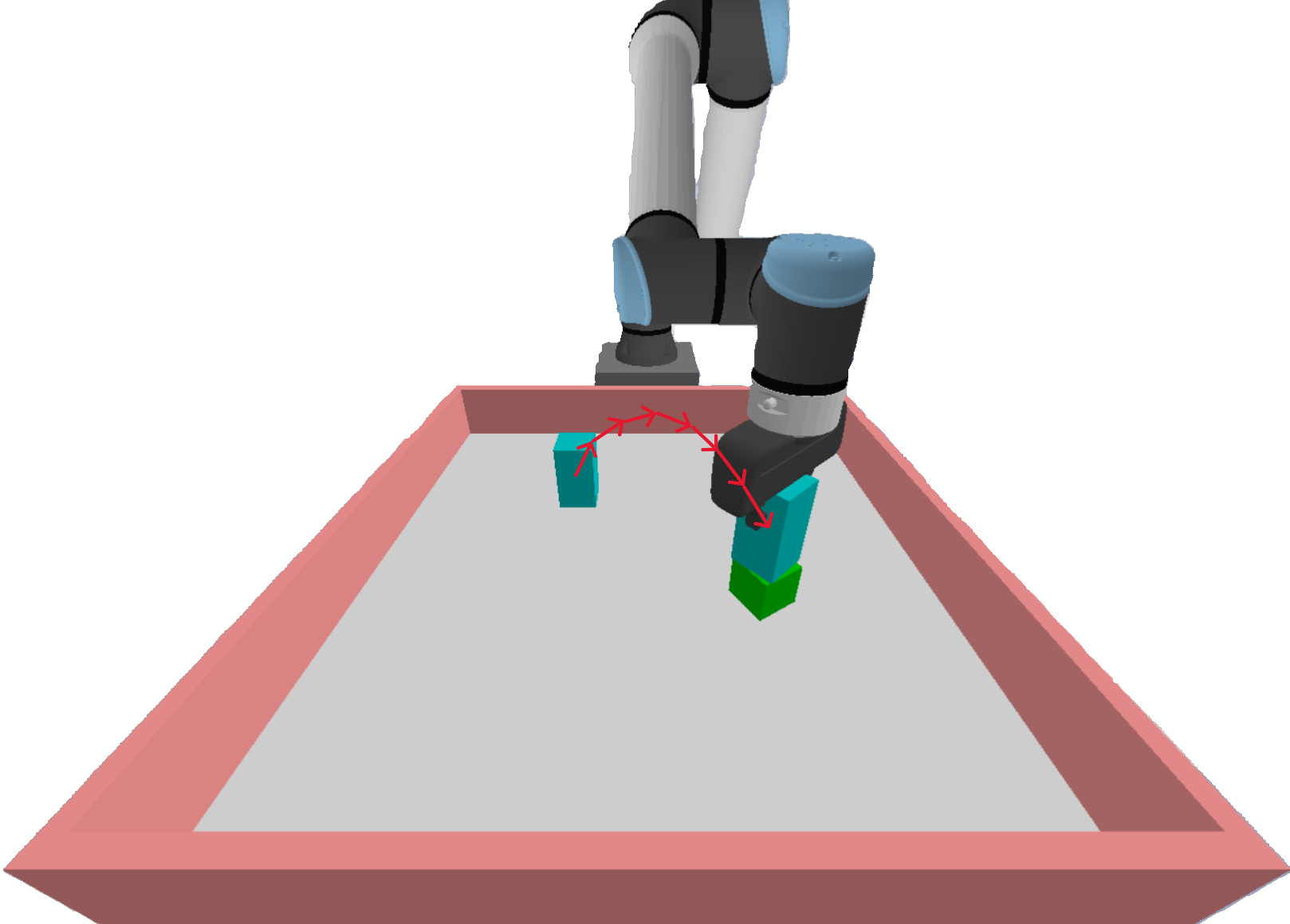}%
    \label{fig:stack}}
    \caption{Visual examples of tasks executed in the PyBullet simulation environment. Arrows in Push (a) and Hit (b) show the trajectory of the object after the action is executed. For Stack (c), arrows show the end-effector trajectory while executing the stacking action.}
\end{figure*}

\textbf{Hit Task.} The second task is a dynamic variation of the Push task. The setup is the same, but the pushing action is applied with \textbf{twice the velocity}, resulting in less predictable object behavior due to possible multiple bouncing off the walls. As we can see in Figures~\ref{fig:hit}, after the vertical cylinder is hit by the robotic arm, it falls over and bounces off from the walls of the table, rolling in the opposite direction of the executed hit action. Here, the tabletop environment and the definitions of $\mathbf{s}$, $\mathbf{a}$, and $\mathbf{e}$ remain identical to the Push task.
\\ \textbf{Stack Task.} In this task, the goal is to place one object on top of another, forming a possible stable stack. This task involves two objects: a `picked' object and a `target' object. As a result, the state includes the positional and orientation information for both objects, doubling the state information compared to the Push and Hit tasks:
\[
\mathbf{s} = [\mathbf{s}_p, \mathbf{s}_t],
\]
where \(\mathbf{s}_p \) and \(\mathbf{s}_t \) are the 9-dimensional states of the picked and target objects, denoted as $o_p$ and $o_t$, respectively. Given the set of six object configurations in the study, this task results in $6\times6=36$ possible object pair combinations. The action is represented by concatenating the one-hot encoding of the picked and the target object:
\[
\mathbf{a} = [\text{onehot}(o_p), \text{onehot}(o_t)].
\]

Similarly, the effect is computed as the change in the objects pre-action and post-action states:

\[
\mathbf{e} = [\mathbf{e}_p, \mathbf{e}_t].
\]

\subsection{Compared Models and Baselines}
Before evaluating the proposed IMTL-EMLP model, we first examine the properties of IMTL-LP, a special case of IMTL-EMLP. We then shift focus to IMTL-EMLP, comparing its performance against both the IMTL-LP model and various baselines. Additionally, we assess different IMTL-EMLP variants with varying energy sensitivity coefficient
$k$, to highlight the trade-off between prediction performance and energy consumption.
\\The baselines used for assessing advantages of our model with LP-based task selection, i.e., IMTL-LP are as follows:
\begin{itemize}
    \item \textbf{SINGLE}: To isolate the impact of shared representation (multi-task learning), our first baseline is a single-task learning model where each task has its own dedicated network for training. The proposed network in Figure \ref{fig:arch} is adapted for single-task learning where each task has its own state projection, encoder, action projection, attention, and decoder modules without any shared parameters between tasks. 
    While it is possible to implement single-task learning using a simpler vanilla MLP for each task, we intentionally adopt the same multi-task learning architecture used for our proposed method. This ensures that the comparison is fair, with both multitask and single-task setups having identical model structures, allowing the differences in performance to be attributed solely to the learning strategy rather than architectural discrepancies.
    \item \textbf{IMTL-RAND}: The second baseline uses the
    shared architecture used in our proposed method, without any modification
    %adaptation, 
    but removes the LP-based task selection mechanism. Instead, tasks are chosen randomly at each training iteration, with each task having an equal chance of being selected. This baseline serves to highlight the impact of our \erhan{task selection (interleaving) } %interleaving 
    strategy by comparing it against a scenario where the interleaving is uniformly random. This baseline %entirely stochastic. It 
    helps isolate the effectiveness of the scheduling algorithm from the benefits of shared parametrization.
    \item  \textbf{BLOCK}: The third and final baseline is implemented using the same shared network architecture but trains tasks in dedicated, uninterrupted blocks rather than interleaving them. To account for any potential effects caused by the order in which tasks are learned, we evaluate all six possible permutations of the three tasks. For each possible order, each task gets a `blocked' learning period before moving to the next one. If there are $R$ training epochs in total, each task gets $R/m$ number of trainings.
    This baseline enables us to assess whether the interleaved approach offers improved learning over all possible blocked learning schedules.
\end{itemize}
For the experiments with the proposed model IMTL-EMLP, we compare it with different variants of it as follows, in addition to the baselines explained above and proposed IMTL-LP model:

\begin{itemize}
    \item \textbf{IMTL-EMLP-K}: To explore how sensitivity coefficient $k$ affects the learning performance and energy usage, we compare IMTL-EMLP models trained with varying $k$ values: IMTL-EMLP-K=0.4, IMTL-EMLP-K=0.7, IMTL-EMLP-K=1, IMTL-EMLP-K=1.2.
\end{itemize}

\subsection{Network Architecture and Initialization}
All models are implemented using the PyTorch \cite{Ansel_PyTorch_2_Faster_2024} framework, using the same set of hyper-parameters across all baseline methods and our proposed model: a learning rate of $0.0001$, a batch size of $100$, a hidden dimension of $4$, and training for $3000$ epochs. Each encoder and decoder module, whether shared or task-specific, consists of two fully connected layers, with ReLU used as the activation function. Optimization is performed using the AdamW optimizer \cite{loshchilov2017decoupled}, with the AMSGrad variant enabled and default weight decay settings. 
To accelerate neural network training, the interactions of the robot with the environment is cached prior to learning so that for each robot interaction the dynamic simulator need not be invoked. An experience cash size of $\approx10\text{K}$ suffice as the tasks used in the experiments have low action and state dimensions.

\begin{figure}[ht]
    \centering
    \includegraphics[width=0.5\textwidth]{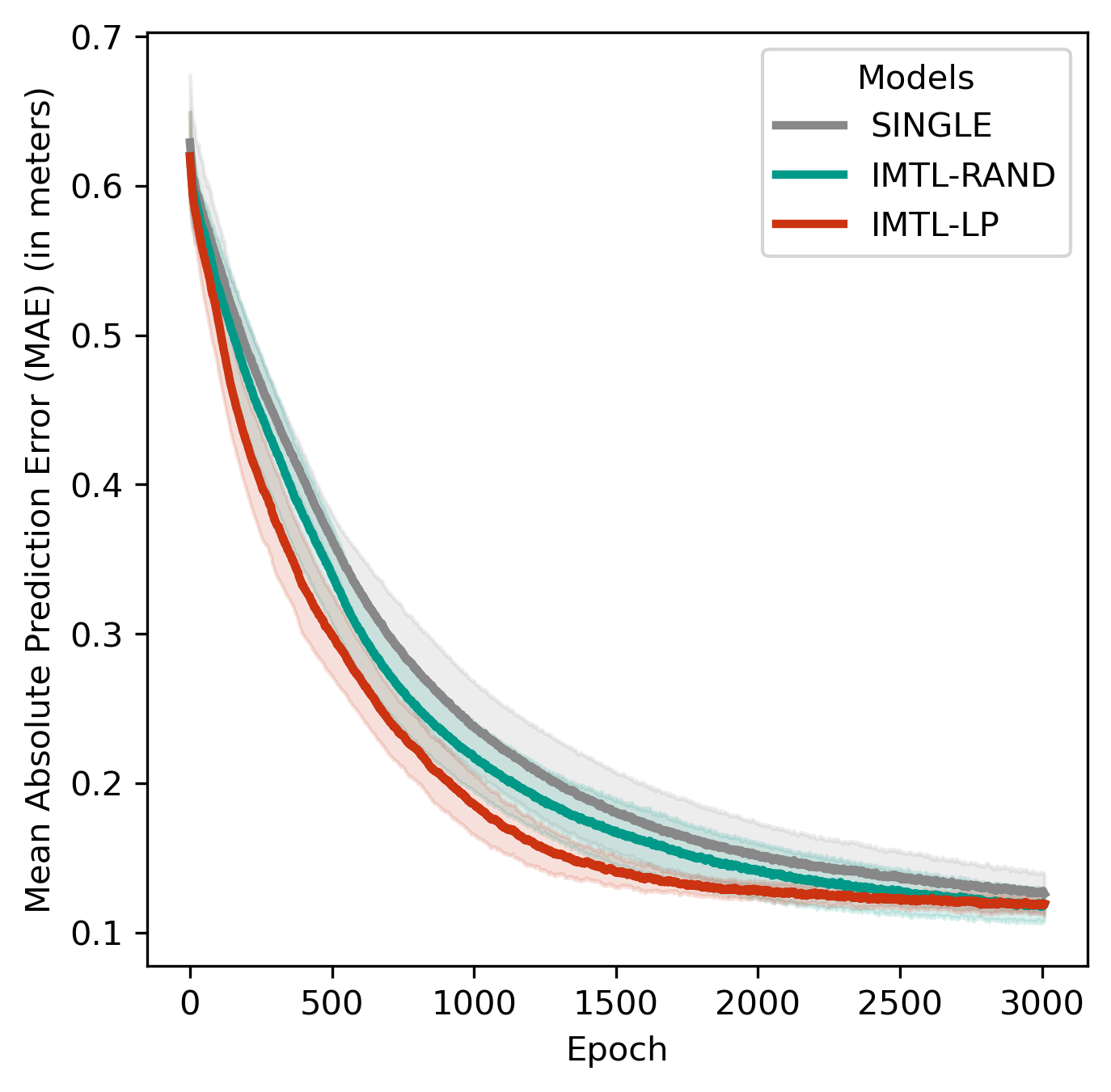}
    \caption{Overall performance of baselines is shown for \erhan{isolated task learning (\textbf{SINGLE}), proposed architecture but random task selection (\textbf{IMTL-RAND}) and proposed architecture with LP-based task selection model (\textbf{IMTL-LP}). As can be seen proposed model (\textbf{IMTL-LP}) converges faster than the baselines.} Even though the random task \erhan{selection} is slightly better than the single task learning, it could not \erhan{surpass learning with LP-based task selection}.}
    \label{fig:overall}
\end{figure}

\subsection{LP-based Interleaved Learning (IMTL-LP) Results}
The initial evaluation compares the proposed model with no energy modulation, i.e., IMTL-LP to  SINGLE and IMTL-RAND baselines. Experiments were conducted using 10 random seeds, and the average task performance results are presented here. As illustrated in Figure \ref{fig:overall}, the proposed model IMTL-LP reduces prediction error faster than the other two baselines. This indicates that the LP-based task selection with inter-task skill transfer produces an interleaved task schedule that
leads to improved overall performance compared to random selection and independent learning with no skill transfer. Furthermore, it is important to note that not only does interleaving enhance performance, learning multiple tasks is also shown to be beneficial, as tasks can mutually support each other and facilitate positive information transfer. This is evidenced by the IMTL-RAND baseline showing improvement compared to the single-task baseline.

\begin{figure}[ht]
  \centering
  \includegraphics[width=\textwidth]{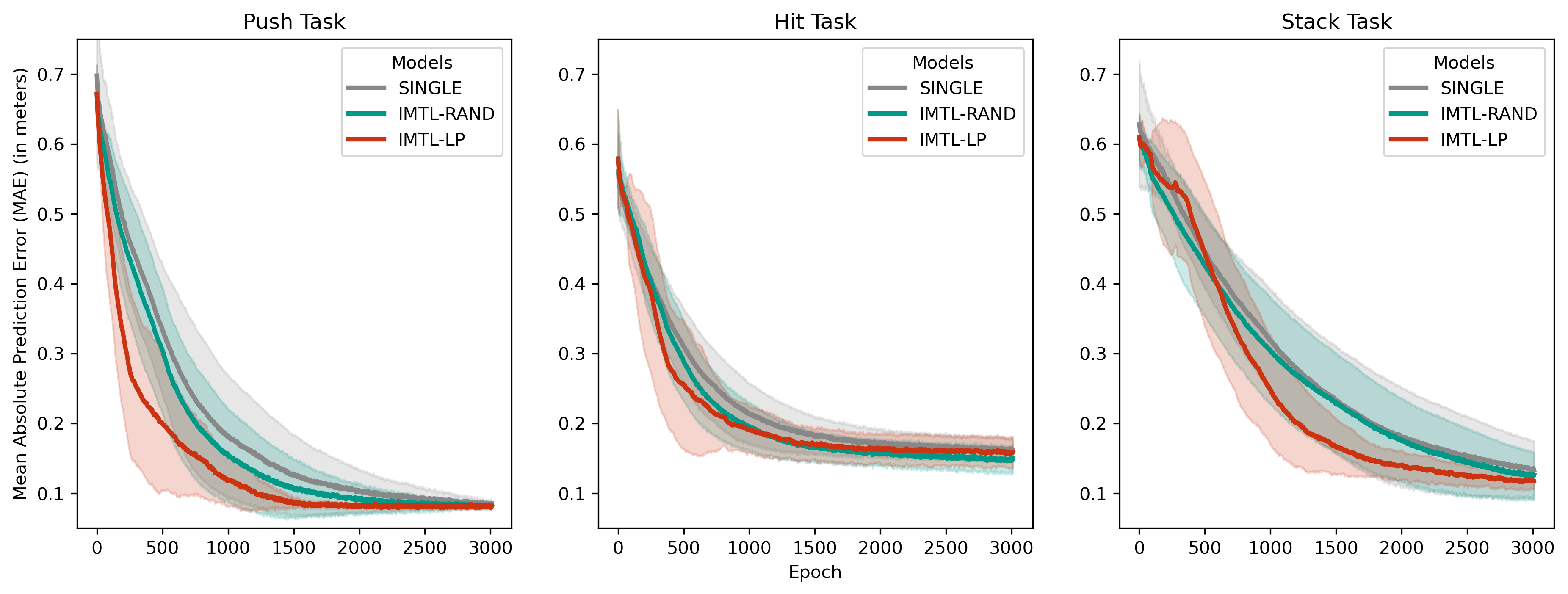}
  \caption{Task-specific model performance is shown. It can be seen that LP-based method surpasses the other baselines on both \textbf{push} and \textbf{stack} tasks, with a higher significance in the latter. However, for the \textbf{hit} task, the improvement was constrained to the learning period around \(\approx 400\)–\(1000\).}
  \label{fig:task-loss}
\end{figure}

\emph{Task-wise Analysis.}
\erhan{To better understand the learning dynamics,} we examine the performance of individual tasks throughout the learning process. As illustrated in Figure \ref{fig:task-loss}, the performance on both the Push and Stack tasks improves when all tasks are learned simultaneously, with a marked enhancement observed when using the proposed model. Notably, the Stack task exhibits a significant performance gap
between our proposed model, IMTL-LP, and the other two baselines.
For the Hit task, as in the other task we see an early reduction in prediction error with IMTL-LP but IMTL-RAND also starts to perform competitive in the second half of training. Table~\ref{tab:inter-count} shows the total iterations allocated for each task due to autonomous task selection. A higher performance can be obtained by IMTL-LP compared to IMTL-RAND in stack task even though it is allocated a shorter training time in total, which indicates that the amount of total training time allocated is not a direct determinant of task learning performance.

\begin{table}[h]
    \centering
    \begin{tabular}{|c|c|c|c|}
    \hline
    Model & Push   &  Hit & Stack \\ \hline \hline
    IMTL-RAND & 1002.5 $\pm 24.4$  & 997.7 $\pm 19.9$  & 1008.8 $\pm 16.8$\\ \hline 
    IMTL-LP & 1525.6 $\pm 124.0$  & 624.8  $\pm 93.2$ & 850.6 $\pm185.6$ \\ \hline
    \end{tabular}
    \caption{The average total number of training iterations received by each task as the result of autonomous task selection in IMTL-LP is compared with uniformly random task selection (IMTL-RAND). The average and standard deviation is over 10 training sessions.}
    \label{tab:inter-count}
\end{table}

\subsection{Effect of Network Complexity on Learning Performance}
\begin{figure}[ht]
    \centering
    \includegraphics[width=0.5\textwidth]{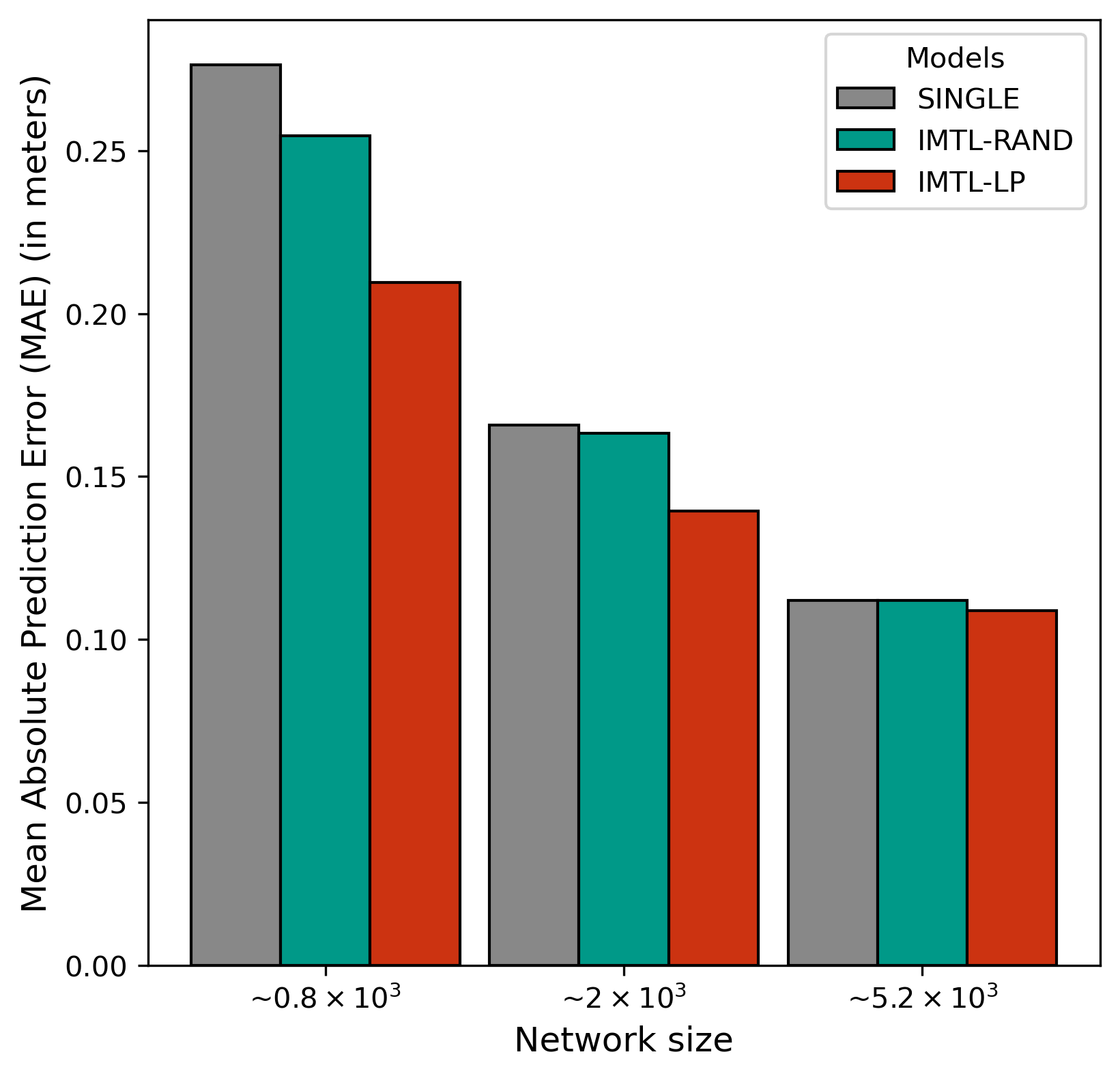}
    \caption{Learning performance of baselines (SINGLE, IMTL-RAND) and our model (IMTL-LP) in different network complexity configurations. Results are taken at mid-training.}
    \label{fig:complexity}
\end{figure}

We analyzed the impact of network complexity on model performance by evaluating SINGLE, IMTL-RAND, and our proposed IMTL-LP (no energy modulation) at three levels of complexity: \textbf{low} (800 parameters), \textbf{medium} (2000 parameters), and \textbf{high} (5200 parameters), as shown in Figure~\ref{fig:complexity}. At low complexity, IMTL-LP significantly outperformed both baselines, highlighting its efficiency in resource-constrained conditions due to our learning progress-based interleaving strategy. At medium complexity, IMTL-LP still outperformed the baselines, although the gap narrowed, suggesting that while SINGLE and IMTL-RAND benefited from additional parameters, IMTL-LP maintained its advantage through better task management. At high complexity, the differences among the three models further decreased, with similar performance across models, indicating that increased capacity partially compensates for limitations in baseline strategies. However, IMTL-LP consistently remained competitive or best performing across all complexity levels.

\subsection{Interleaved vs. Blocked Learning}
\begin{figure*}[!t]
    \centering
    \includegraphics[width=\textwidth]{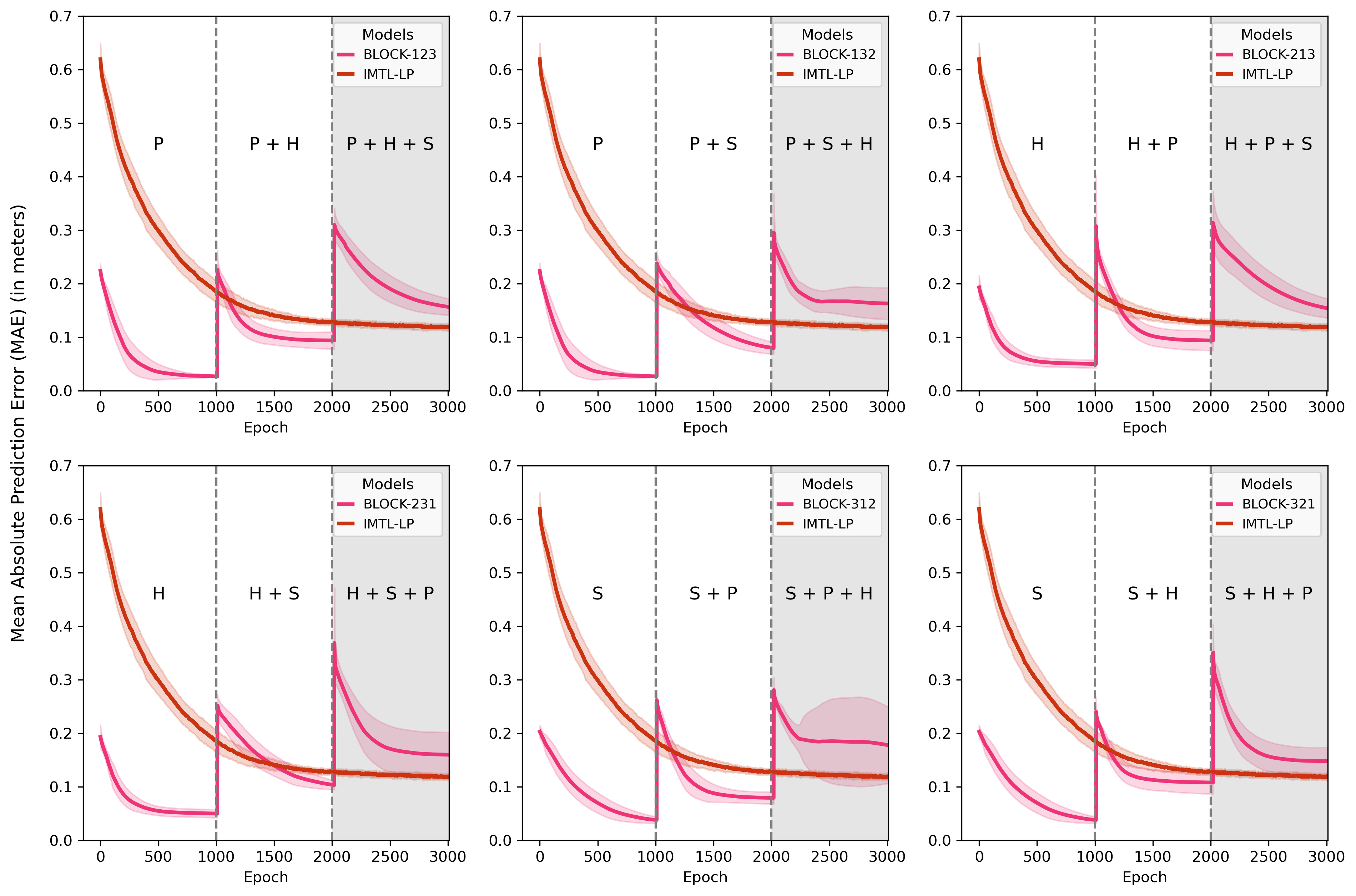}
    \caption{Interleaved learning versus blocked learning prediction error plots for each task order combination. Each figure has three regions divided by vertical dashed lines, indicating where the BLOCK model switches to the next task in the order. P stands for Push, H stands for Hit and S stands for Stack task. Rightmost region in each figure shows where both models, BLOCK and IMTL-LP, have seen all the tasks to be learned. In all the task orders, IMTL-LP surpasses block learning.}
    \label{fig:blockedvsimtl}
\end{figure*}

Blocked learning is a training paradigm in which tasks are presented sequentially in distinct, uninterrupted intervals. It can be questioned that the blocked learning scheme may match or surpass the performance observed with interleaved learning when a certain task order is used. To test for this, we  conducted a detailed comparison of interleaved learning and blocked learning with a neural network suitable for multi-task learning across the tasks of Push, Hit, and Stack introduced in Section \ref{sec:method}. 

Given the three tasks, there are six possible training orders (e.g., Push $\to$  Hit $\to$ Stack or Push $\to$ Stack $\to$ Hit, etc.). \erhan{To see whether any order can perform similar or bettered that interleaved learning,} we trained the BLOCK model across all six task orders (task IDs are denoted in the order as -XXX, e.g., BLOCK-123), while the IMTL-LP model remained unchanged across these configurations, as it does not depend on the task order (see Table~\ref{tab:inter-count} for per task training count of IMTL-LP model). 

Figure~\ref{fig:blockedvsimtl} shows the performance of the IMTL-LP model and the BLOCK model for each of the six task orders. It can be seen that the proposed model, IMTL-LP consistently outperforms the BLOCK model across all task orders. No matter the order in which the tasks are trained, our IMTL-LP model always surpasses the blocked learning method, which shows that interleaved learning prevents catastrophic forgetting of learned information by switching the context in periods, thus being able to recall faster. On the other hand, blocked learning as expected suffers from catastrophic forgetting as evidenced by the error spikes at the task switch boundaries. Additionally, the variability in performance across different task orders for the BLOCK model indicates its sensitivity to the order of training.

\begin{figure}[ht]
    \centering
    \includegraphics[width=0.5\textwidth]{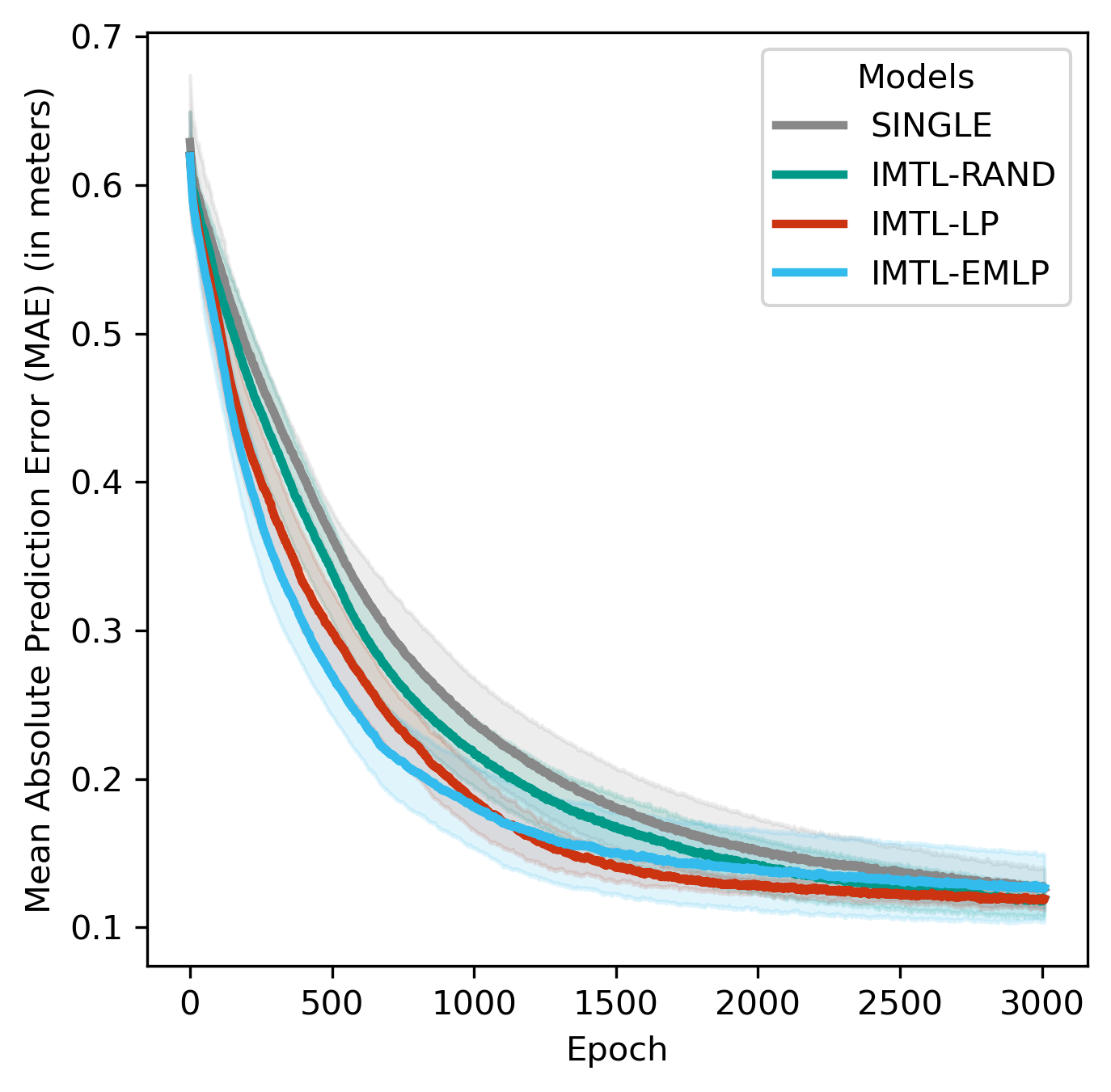}
    \caption{Comparison of baselines and LP-based model (IMTL-LP) to the EMLP-based model (IMTL-EMLP) where the sensitivity constant is set to $k=1$.}
    \label{fig:lpe-loss}
\end{figure}

\subsection{Modulation of LP with Computational Energy (IMTL-EMLP)}
In this section, inspired form human brain's tendency to reduce computational cost, we explore discounting the learning progress of the tasks by the neural cost they incur during learning. For this, we first test IMTL-EMLP  that uses task selection according to  $e^{k\cdot\tilde{LP}^{\mathcal{T}_i}}/\tilde{EC}^{\mathcal{T}_i}$ where the relative importance of LP set as $k=1$ (see Equation~\ref{eq:energydiscount} ) and compare it to IMTL-LP and other baselines. Then  we run experiments with different values of $k$ and compare the resulting learning performances in terms of prediction error and energy expenditure. Figure~\ref{fig:lpe-loss} presents the overall learning performance of our proposed interleaved learning (IMTL-LP and IMTL-EMLP)  with the baseline models (SINGLE and IMTL-RAND), in terms of prediction performance. We can clearly say that our energy-modulated model, IMTL-EMLP, shows the best performance until 1000 epochs, being slightly better than our model with no energy modulation (IMTL-LP). After that, both of them show similar performance.
The IMTL-EMLP model also outperforms both SINGLE and IMTL-RAND baselines, indicating that incorporating energy awareness does not nullify the gains of our architecture. The SINGLE model exhibits the weakest performance overall, reaffirming that task isolation limits skill learning efficiency. Meanwhile, IMTL-RAND shows modest improvements over SINGLE, highlighting that even random task interleaving can offer some benefit, yet remains clearly inferior to structured LP-based scheduling. Taken together, these results confirm that both learning progress and energy-aware arbitration strategies provide meaningful gains in training efficiency and final task performance, with IMTL-LP and IMTL-EMLP variants, converging faster and reaching lower prediction errors than all baselines.

\begin{figure*}[t]
    \captionsetup[subfloat]{labelfont=scriptsize,textfont=scriptsize} 
    \centering
    \subfloat[Training performances of the models.]{\includegraphics[width=0.5\textwidth]{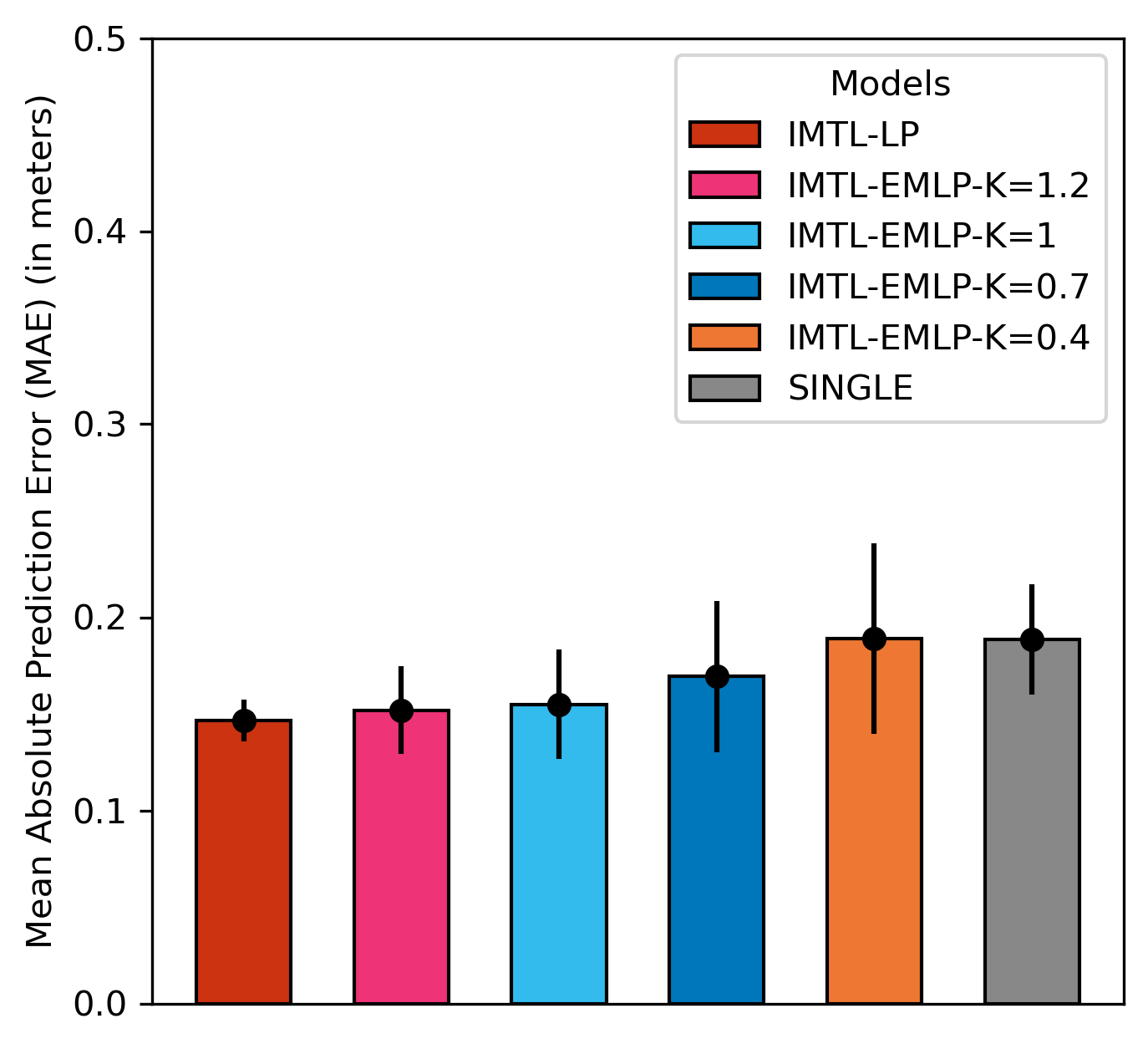}%
    \label{fig:k-loss}}
    \hfill
    \subfloat[Energy consumption of the models.]{\includegraphics[width=0.5\textwidth]{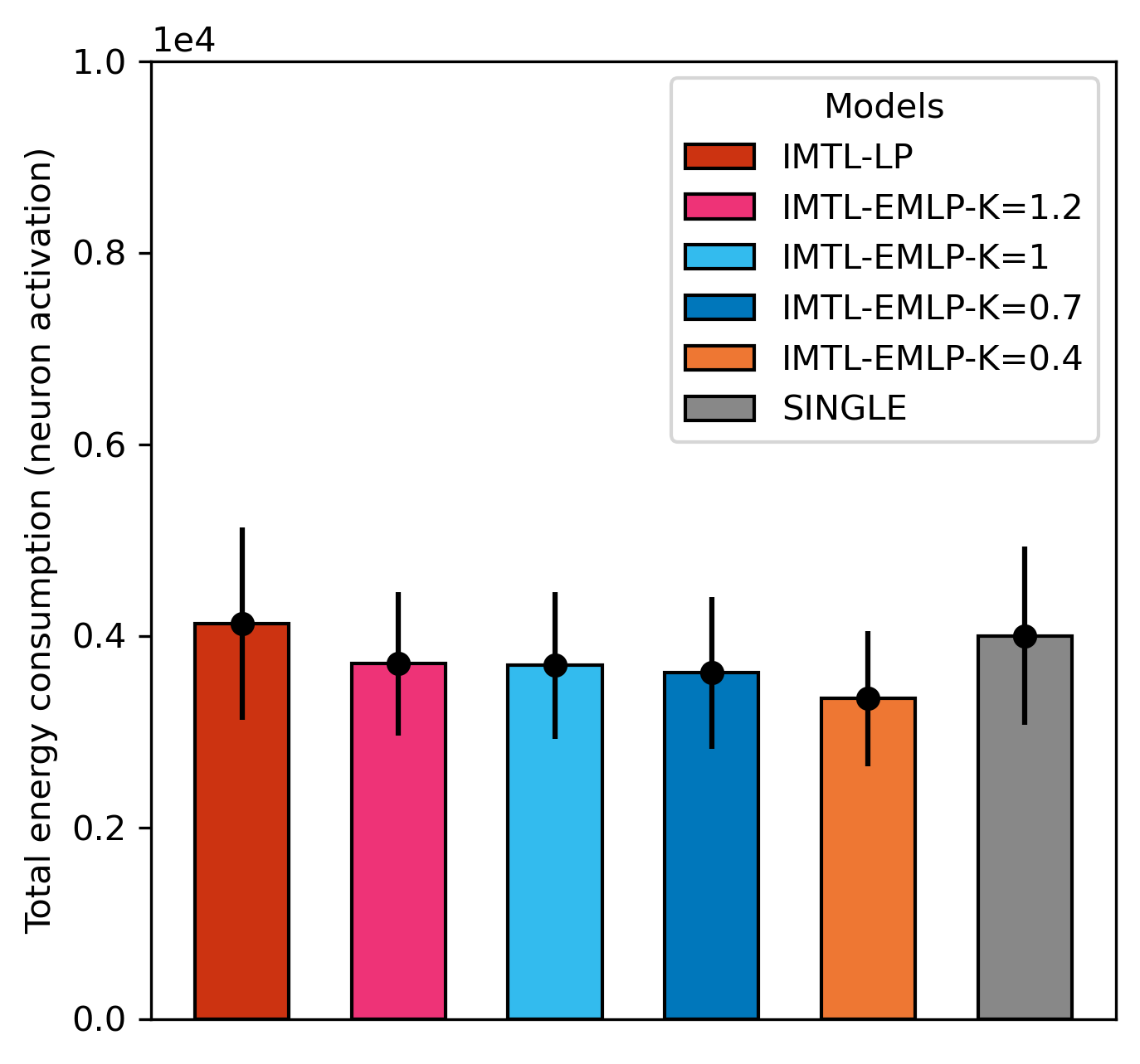}%
    \label{fig:k-energy}}
    \caption{Effect of varying the energy sensitivity coefficient $k$ in the IMTL-EMLP model on (a) prediction performance and (b) total energy consumption, at the midpoint of training. Higher values of $k$ prioritize learning progress over energy efficiency, leading to lower prediction error but increased energy usage, while lower $k$ values reduce energy consumption at the cost of predictive accuracy.}

\end{figure*}

To further evaluate the effect of EMLP-based task scheduling, we investigate the performance of the IMTL-EMLP model under varying values of the energy sensitivity coefficient 
$k$, and compare it to IMTL-LP and SINGLE baselines. As shown in Figure~\ref{fig:k-loss}, which reports the MAE at the midpoint of the training ($\sim1500$), IMTL-LP achieves the lowest prediction error overall, confirming the strong effectiveness of LP-based task arbitration. Then, IMTL-EMLP-K=1.2 closely follows, demonstrating that modest energy modulation can preserve much of the learning performance. As $k$ decreases, the prediction performance gradually degrades, indicating that overly prioritizing energy conservation (i.e., smaller $k$) can hinder learning effectiveness. On the other hand, Figure~\ref{fig:k-energy} shows total energy consumption measured in terms of cumulative neuron activations. Here, we observe the opposite trend: energy usage decreases steadily with smaller 
$k$, with IMTL-EMLP-K=0.4 consuming the least energy overall, almost the same as the SINGLE baseline. While IMTL-LP achieves the best predictive performance, it is also the most energy-intensive. Notably, IMTL-EMLP-K=1.2 and K=1 provide a favorable trade-off, offering reductions in energy consumption with only a small sacrifice in prediction accuracy. This tunable behavior highlights the flexibility of the IMTL-EMLP framework, enabling agents to balance performance and energy efficiency according to task demands and environmental constraints.

\subsection{Ablation Experiments}
In order to reveal the individual affects of the self-attention and the task flag introduced in the proposed multitask learning architecture, a series of ablation experiments are conducted. In addition, task skill transfer ablations are conducted to investigate the skill transfer relations formed due to interleaved learning.
\subsubsection{Self-Attention  and the Task Flag Ablations}
\begin{figure}[t]
    \centering
    \includegraphics[width=0.5\columnwidth]{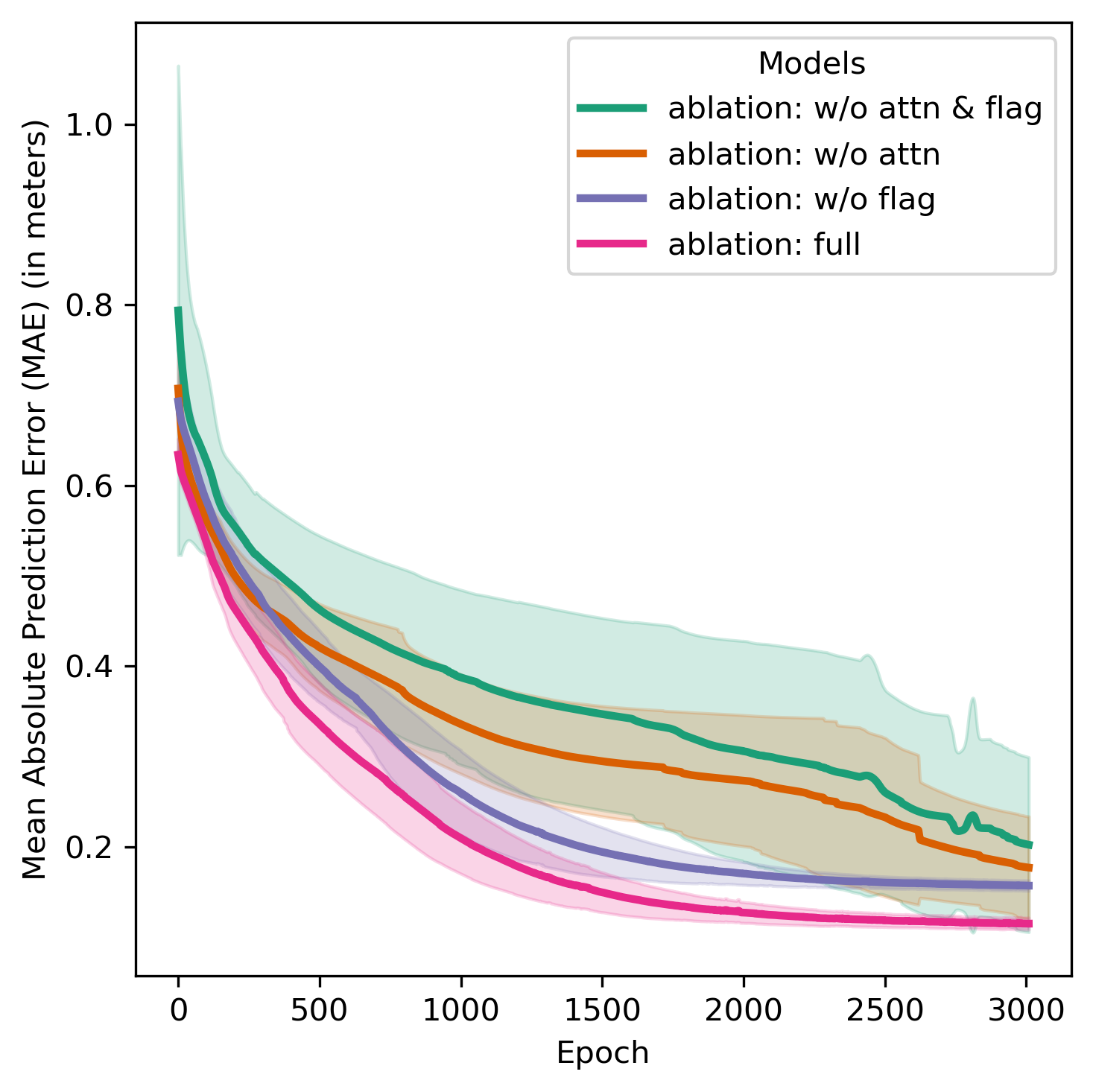}
    \caption{Contributions of the shared attention and flag bit from the proposed multi-task learning architecture to the model performances. Neither attention nor flag alone can surpass the convergence speed of our proposed model, which combines them.}
    \label{fig:ablation-attention}
\end{figure}
The individual contributions of the attention layer and the flag bit in our proposed model are evaluated using four versions of the model\erhan{: full model, without task flag, without attention layer,  without flag and attention layer.}  Each model was trained using 10 random seeds. Below, we describe the four ablation settings and provide a detailed analysis of their performance: 
\\\emph{Full model.} The proposed model including a shared attention layer that process the task representations with the flag bit.
\\ \emph{Model w/o task flag.} This variation retains the shared attention layer but removes the flag bits. Task representations are combined and passed to the shared attention layer without explicit differentiation using flags.
\\ \emph{Model w/o attention layer.} In this model, the flag bits are concatenated to the task representations, same as the original matrix $Z$ from Equation~\eqref{eq:matrix}, but instead of the shared attention layer, $Z$ is forwarded directly to task decoder of the current training task.
\\ \emph{Model w/o attention layer and flag.} This configuration combines the two ablations thus only concatenated task representations are relayed to the task decoder of the current training task.

The results of the ablation experiments are shown together in Figure~\ref{fig:ablation-attention} as prediction error versus training epoch curves. \erhan{We can deduce the following from the ablation experiments.
The model without both the attention mechanism and the task-identifying flag bit (Figure~\ref{fig:ablation-attention}, w/o attn\&flag) showed the poorest performance, indicating that the absence of both components severely hinders the model’s ability to share information across tasks and distinguish between them. Introducing only the flag bit (Figure~\ref{fig:ablation-attention}, w/o attn) improved the results, as it allowed for explicit task differentiation, though the lack of attention limited the model’s capacity to capture dynamic task relationships. Conversely, using the attention mechanism without the flag (Figure~\ref{fig:ablation-attention}, w/o flag ) led to better performance, as the model could exploit shared information across tasks despite the absence of explicit task cues. The full model, incorporating both the attention layer and the flag bit, consistently outperformed all other configurations, demonstrating that the two mechanisms are complementary and together provide significant benefits for multi-task learning from the early stages of training.
}

\subsubsection{Object-wise Skill Transfer Analysis}

\begin{figure*}[!t]
    \captionsetup[subfloat]{labelfont=scriptsize,textfont=scriptsize} 
    \centering
    \subfloat[Task-wise performance degradation with per-object data, where rows are target tasks and columns are source tasks respectively.] 
    {\includegraphics[width=\textwidth]{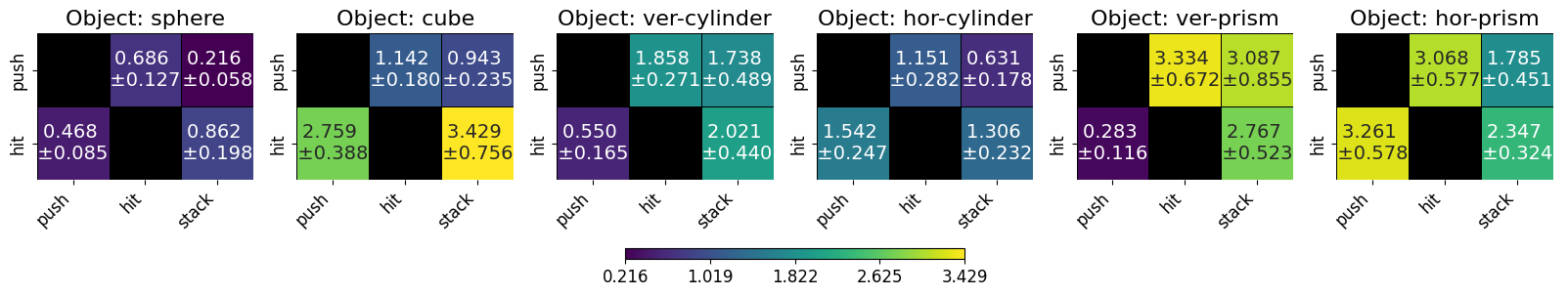}%
    \label{fig:one-obj-transfer}}
    \vspace{0.2cm}
    \subfloat[Stack task performance degradation with per-target object data, where rows are picked objects by the robotic arm and columns are source tasks respectively.] 
    {\includegraphics[width=\textwidth]{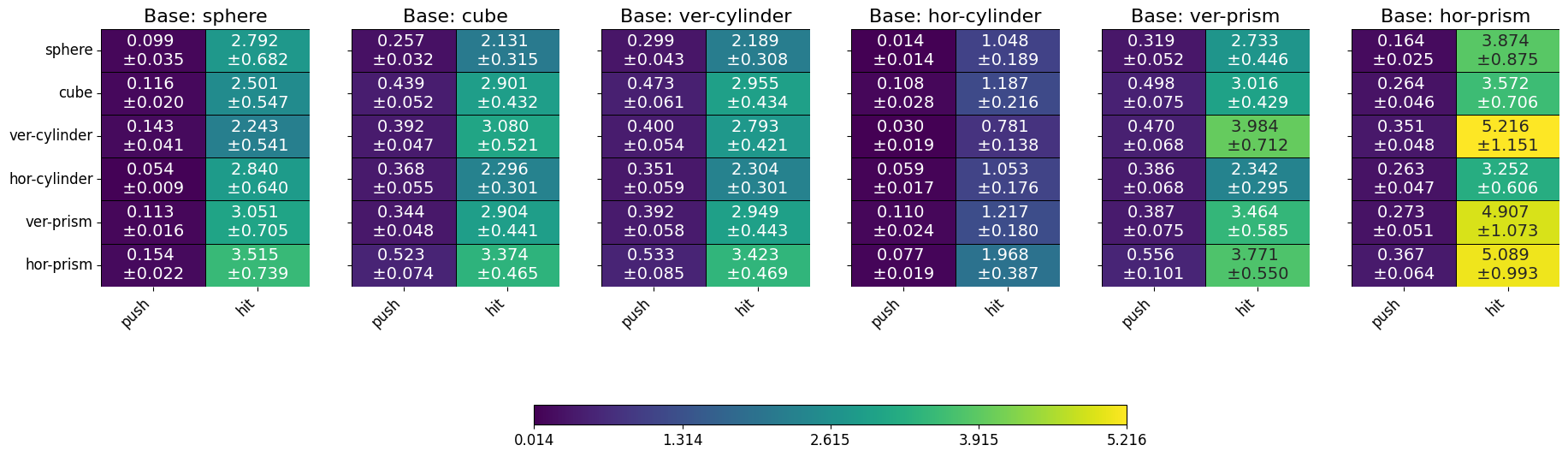}%
    \label{fig:two-obj-transfer}}
    \caption{\erhan{Impact of source-to-target task ablations on target task performance across objects.} Each sub-figure corresponds to a specific object and visualizes the performance degradation (increase in loss) for target tasks (columns) when the representation of a particular source task (rows) is removed from the key and query inputs of the shared multi-head attention module. This experiment reveals the dependency of each target task on the shared representation of other source tasks, highlighting the inter-task transfer and interference dynamics in the multi-effect prediction framework.}
    \label{fig:skilltransfer}
\end{figure*}
In this section, we present skill transfer among tasks that were quantified by ablating task-to-task projections in the shared attention layer after training is completed. To assess a source task's contribution to a target task, we define the loss obtained on the target task when all task representations are intact in the key/value matrices as $\mathcal{L}_{\mathrm{full}}$ and use $\mathcal{L}_{\mathrm{ablate}}$ to denote the loss obtained after zeroing out the representation of the source task. If the source task is irrelevant for the target task, a large increase in the loss is not expected. On the other hand, if the source task is utilized, then the loss should increase. With this logic, we define $\Delta\mathcal{L}=\mathcal{L}_{\mathrm{ablate}}-\mathcal{L}_{\mathrm{full}}$ and interpret the results based on this metric. For example, given a (source, target) task pair, a positive $\Delta\mathcal{L}$ indicates positive transfer from the source task to the target task. 
\erhan{For skill transfer analysis, we focused on our IMTL-LP model and repeated training \& testing 10 times with random seeds. The average impact of the ablations are measured by  $\Delta\mathcal{L}$ and its standard deviation and used for numeric and visual representation (see Figure~\ref{fig:skilltransfer}).} 

The skill transfer analysis is carried out  per-object basis; so, the reported average $\Delta\mathcal{L}$ for a source and target task pair
indicates the skill transfer between the tasks when a particular object is being handled. 
\\First, we present the results for single-object interactions (Push and Hit), and then look at two-object interactions (Stack). Figure~\ref{fig:one-obj-transfer} shows skill transfer levels between source tasks (columns) and target tasks (rows) as a color coded matrix.  During interaction with \emph{sphere}, skill transfer was lower than interacting with the other objects. With \emph{sphere}, the highest $\Delta\mathcal{L}$ is observed in $Stack \rightarrow Hit$ case with $0.86$, whereas the lowest is observed in $Stack \rightarrow Push$ ($0.22$). Hit seems to better exploit Stack compared to Push suggesting that richer dynamics of multiple reflections from the walls (due to hit action) and the falls during unstable stacking shares shares certain computational similarities. Supporting this interpretation, in the \emph{horizontal cylinder} a similar pattern can be observed where Stack benefits Hit more than Push. The common \emph{rollability} affordance of sphere (2D) and the horizontal cylinder  (1D) seem to limit the skill transfer among tasks when the interacted object is rollable. This may be explained by the fact that the behavior variety one can observe from rollable objects is higher compared to non-rollable objects, thereby making consistent skill transfer difficult. One difference in the sphere and the horizontal cylinder is the relative high skill transfer observed in  $Push \rightarrow Hit$. This can be understood by considering that the push action is applied from arbitrary directions with many of them not inducing a rolling behavior. In those cases, the behavior of the object is closer to a stable object such as cube or horizontal-prism and thus the transfer level is somewhat shifted toward higher levels as in the stable objects (see horizontal prism and cube $Push \rightarrow Hit$ skill transfers). The effects of affordance and orientation can be seen in the skill transfer level of the cylinder and the prism: the  $Push \rightarrow Hit$ transfer is limited when the objects are vertical (cylinder:~$0.55$; prism:~$0.28$), whereas the transfer is enhanced when they are placed horizontally (cylinder:~$1.55$; prism:~$3.26$). It seems that Hit action becomes beneficial to Push when objects are oriented vertical as opposed to horizontal so that they are knock-able. In the cylinder case, altering the orientation from horizontal to vertical causes a transfer change of $1.15$ to $1.85$; similarly in the prism case, a change of $3.07$ to $3.33$ is observed.

Moving on to the two-object interactions in Figure~\ref{fig:two-obj-transfer} for the Stack task, each matrix is titled  with the base object on to which the picked object is placed. Rows show the picked object and columns show the source tasks (Push and Hit). The most prominent observation is that Hit task is strongly utilized by the Stack task compared to the Push task that have a non-negative small contribution.
The high transfers level ($\Delta\mathcal{L} \approx 3.7 - 5.2$) from Hit task are observed when the object the base object stays intact, as in when the pair can be stacked successfully. The transfer levels when the base object is a sphere or a horizontal cylinder follow a similar pattern to that of one-object interactions. Their rollability affordances in 2D and 1D, respectively, cause them to behave differently depending on whether they receive a low impact force (push) or high impact force (hit). Cube or vertical cylinder as the base object show almost the same skill transfer pattern. 

Considering these object-wise analyzes, along with the task-specific learning performance figures (see Figure~\ref{fig:task-loss}), we see that Push task learns the effect prediction problem mostly on its own during the early stages of learning, while Hit and Stack tasks co-learn and benefit from each other throughout the training. It is intriguing to see that even though the task selection and learning performance is affected by the random initializations, the standard deviation of the effects of the ablations is about 14\%. This indicates that although different skill transfer patterns are possible, the current results reflect a general pattern for the tasks addressed. Overall, we can conclude that, our task agnostic task selection mechanism, albeit being not geared to elicit skill transfer, sculpts an emergent transfer landscape that leads to faster and improved learning.

\section{Conclusion}\label{sec:conc}
This paper introduced a biologically inspired interleaved multi-task learning framework that selects tasks dynamically based on learning progress and neural computational energy cost. Motivated by how humans interleave tasks and regulate cognitive effort, our method, IMTL-LP, prioritizes tasks that exhibit positive learning progress, while the energy-aware version, IMTL-EMLP, considers learning efficiency by discounting  LP by the energy it takes to achieve it, shifting the preference to the tasks with lower energy consumption. The learning architecture developed follows a shared encoder-decoder structure equipped with bi-directional skill transfer mechanisms that can learn multiple tasks with dynamic task engagement.

The developed framework is evaluated on a robotic manipulation scenario where the robot is tasked with learning the effects of its actions in different task environments. The experimental results demonstrate that LP-based task arbitration not only improves overall learning performance and convergence speed compared to baseline strategies like random interleaving, single-task learning, and blocked training, but also facilitates beneficial knowledge transfer across tasks. The results indicate that it is not only the training time allocated for the tasks that affect the learning performance, but  also the task scheduling is critical, which is autonomously generated by our model. Moreover, the inclusion of energy-awareness in the task selection process enables a tunable trade-off between learning accuracy and resource efficiency, which is especially valuable in energy-constrained settings. Our ablation studies further validate the complementary roles of the attention mechanism and task flag bits in enhancing multi-task learning dynamics.

While the model is tested in a controlled simulation environment with a limited number of tasks, the results point to broader implications for continuous and sustainable learning systems. By drawing from principles of intrinsic motivation and neural efficiency, the proposed framework contributes a novel perspective on how task interleaving and energy-awareness can be adopted in artificial agents. Future work may explore extensions to real-world robotic platforms, integration with reinforcement learning settings, or scaling to larger task sets with more diverse dynamics.

\section*{Acknowledgements}
\erhan{This work was supported by the Japan Society for the Promotion of Science KAKENHI Grant Number JP23K24926. Additional support was provided by the project JPNP16007 commissioned by the New Energy and Industrial Technology Development Organization (NEDO) and Japan Society for the Promotion of Science KAKENHI Grant Number JP25H01236 and by the European Union under the INVERSE project (101136067).}

\newpage
\appendix
\setcounter{table}{0}
\setcounter{figure}{0}
\section{Task Settings and Hyperparameters}
\begin{table}[h!]
    \centering
    \begin{subtable}[t]{0.474\textwidth}
        \centering
        \resizebox{\linewidth}{!}{%
        \begin{tabular}{|c|c|c|c|}
            \hline
            Module  & Layer & In size & Out size \\ \hline \hline
            \rule{0pt}{3ex}            $P^{\mathcal{T}_i}_{\text{state}} : \mathbb{R}^{d_s^{\mathcal{T}_i}} \rightarrow \mathbb{R}^{d_s}$  & FC             & 9/9/18                   & 4                    \\ \hline
            \multirow{2}{*}{$F : \mathbb{R}^{d_s} \rightarrow \mathbb{R}^{d_h}$} & FC+ReLU      & 4                         & 4 \\
                               & FC+ReLU        & 4                         & 4 \\ \hline
            \multirow{2}{*}{$f^{\mathcal{T}_i} : \mathbb{R}^{d_h} \rightarrow \mathbb{R}^{d_r}$ } & FC+ReLU & 4                       & 4  \\
                                        & FC+ReLU & 4                       & 2   \\ \hline
            $MHA: \mathbb{R}^{d_r}  \rightarrow \mathbb{R}^{d_r}$ & MHA                   & 2                  & 2                    \\ \hline \rule{0pt}{3ex}
            $P^{\mathcal{T}_i}_{action} : \mathbb{R}^{d_a^{\mathcal{T}_i}} \rightarrow \mathbb{R}^{d_a}$ & FC             & 8/8/12                  & 1                    \\ \hline
            \multirow{4}{*}{$g^{\mathcal{T}_i} : \mathbb{R}^{d_r+d_a} \rightarrow \mathbb{R}^{d_e^{\mathcal{T}_i}}$ } & FC+ReLU & 3                       & 4  \\
                                        & FC+ReLU & 4                       & 4  \\
                                        & FC+ReLU & 4                       & 4  \\
                                        & FC+ReLU & 4                       & 9/9/18   \\ \hline
        \end{tabular}
        }
        \caption{Input and output dimensions for the single-task learning models.}
    \end{subtable}\hfill   
    \begin{subtable}[t]{0.506\textwidth}
        \centering
        \resizebox{\linewidth}{!}{%
        \begin{tabular}{|c|c|c|c|}
            \hline
            Module  & Layer & In size & Out size \\ \hline \hline
            \rule{0pt}{3ex}            $P^{\mathcal{T}_i}_{\text{state}} : \mathbb{R}^{d_s^{\mathcal{T}_i}} \rightarrow \mathbb{R}^{d_s}$  & FC             & 9/9/18                   & 6                    \\ \hline
            \multirow{2}{*}{$F : \mathbb{R}^{d_s} \rightarrow \mathbb{R}^{d_h}$} & FC+ReLU      & 6                         & 6 \\
                               & FC+ReLU        & 6                         & 4 \\ \hline
            \multirow{2}{*}{$f^{\mathcal{T}_i} : \mathbb{R}^{d_h} \rightarrow \mathbb{R}^{d_r}$ } & FC+ReLU & 4                       & 4  \\
                                        & FC+ReLU & 4                       & 2   \\ \hline
            $MHA : \mathbb{R}^{d_r+1} \rightarrow \mathbb{R}^{d_r+1}$ & MHA                   & 3                  & 3                    \\ \hline \rule{0pt}{3ex}
            $P^{\mathcal{T}_i}_{action} : \mathbb{R}^{d_a^{\mathcal{T}_i}} \rightarrow \mathbb{R}^{d_a}$ & FC             & 8/8/12                  & 1                    \\ \hline
            \multirow{4}{*}{$g^{\mathcal{T}_i} : \mathbb{R}^{d_r+d_a+1} \rightarrow \mathbb{R}^{d_e^{\mathcal{T}_i}}$ } & FC+ReLU & 4                       & 4  \\
                                        & FC+ReLU & 4                       & 4  \\
                                        & FC+ReLU & 4                       & 4  \\
                                        & FC+ReLU & 4                       & 9/9/18   \\ \hline
        \end{tabular}
        }
        \caption{Input and output dimensions for the multi-task learning models.}
    \end{subtable}
    \caption{Network architecture details for both single and multi-task learning models. Number of neurons in $F$ in single-task learning is increased for the multi-task learning model, to ensure both architectures have the same number of parameters overall.}
    \label{tab:param}
\end{table}
\newpage
\section{IMTL-LP Task Selection Regimes}
\begin{figure*}[ht!]
    \captionsetup[subfloat]{labelfont=scriptsize,textfont=scriptsize} 
    \centering
    \subfloat{\includegraphics[width=0.8\linewidth]{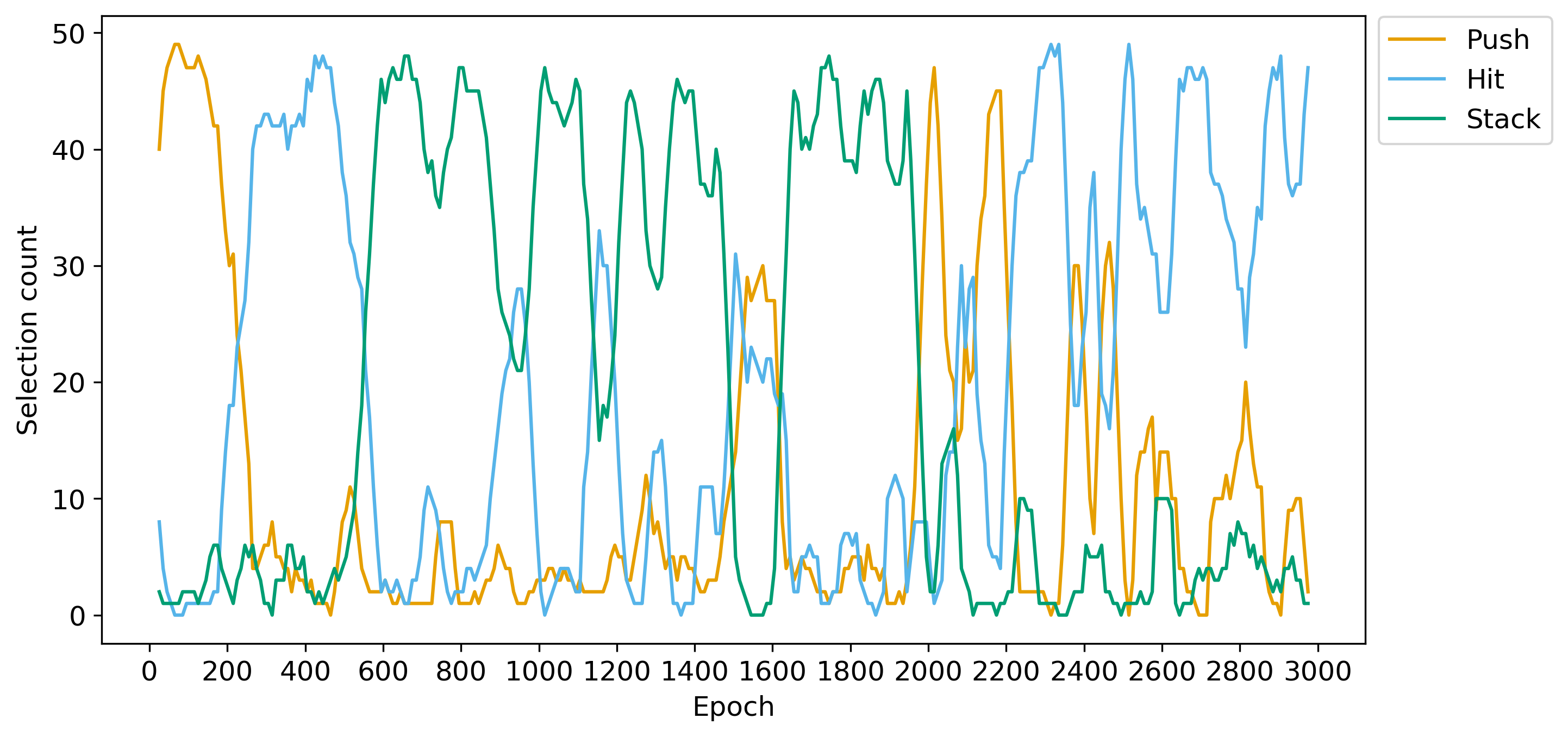}%
    \label{fig:s0}}
    \vfill
    \subfloat{\includegraphics[width=0.8\linewidth]{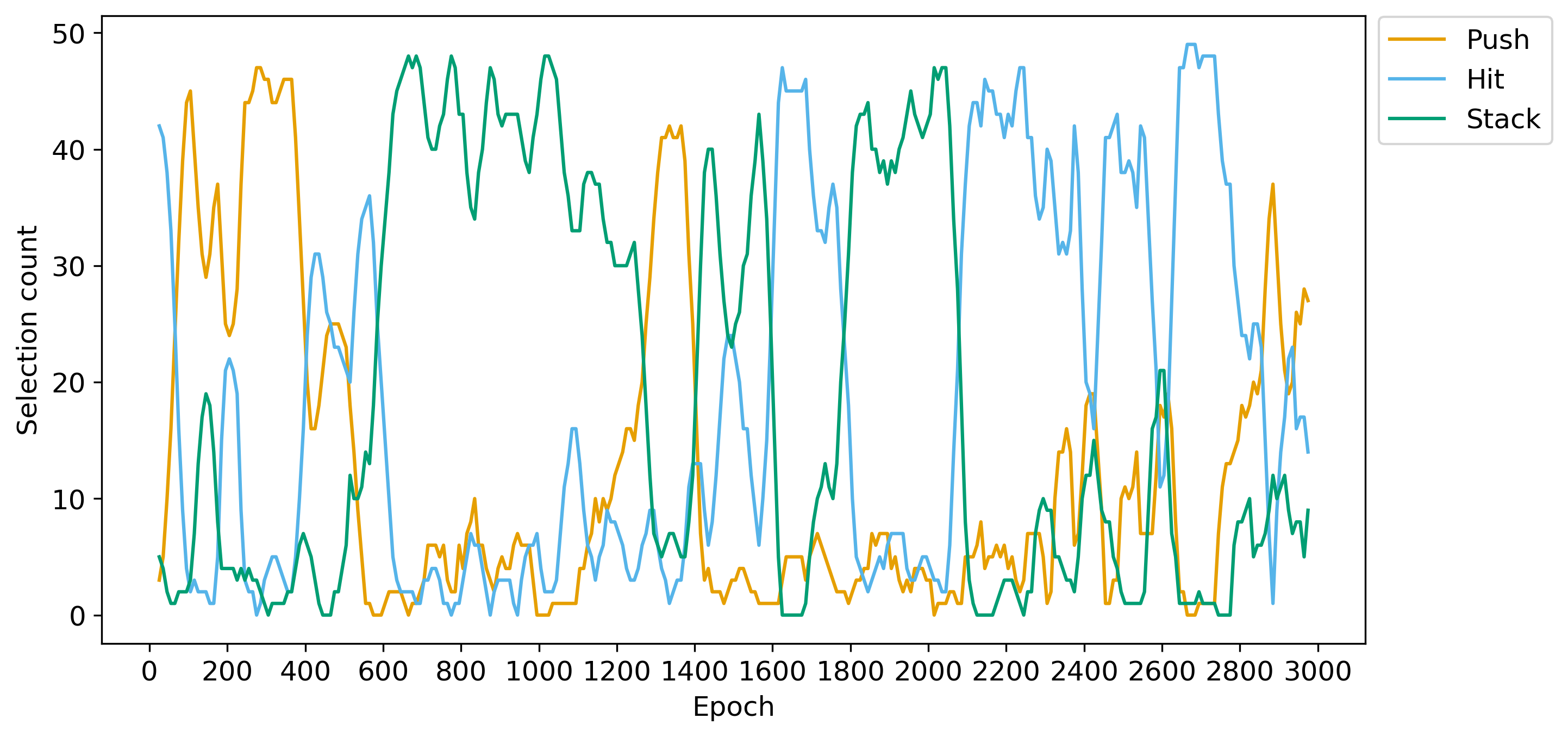}%
    \label{fig:s1}}
    \vfill
    \subfloat{\includegraphics[width=0.8\linewidth]{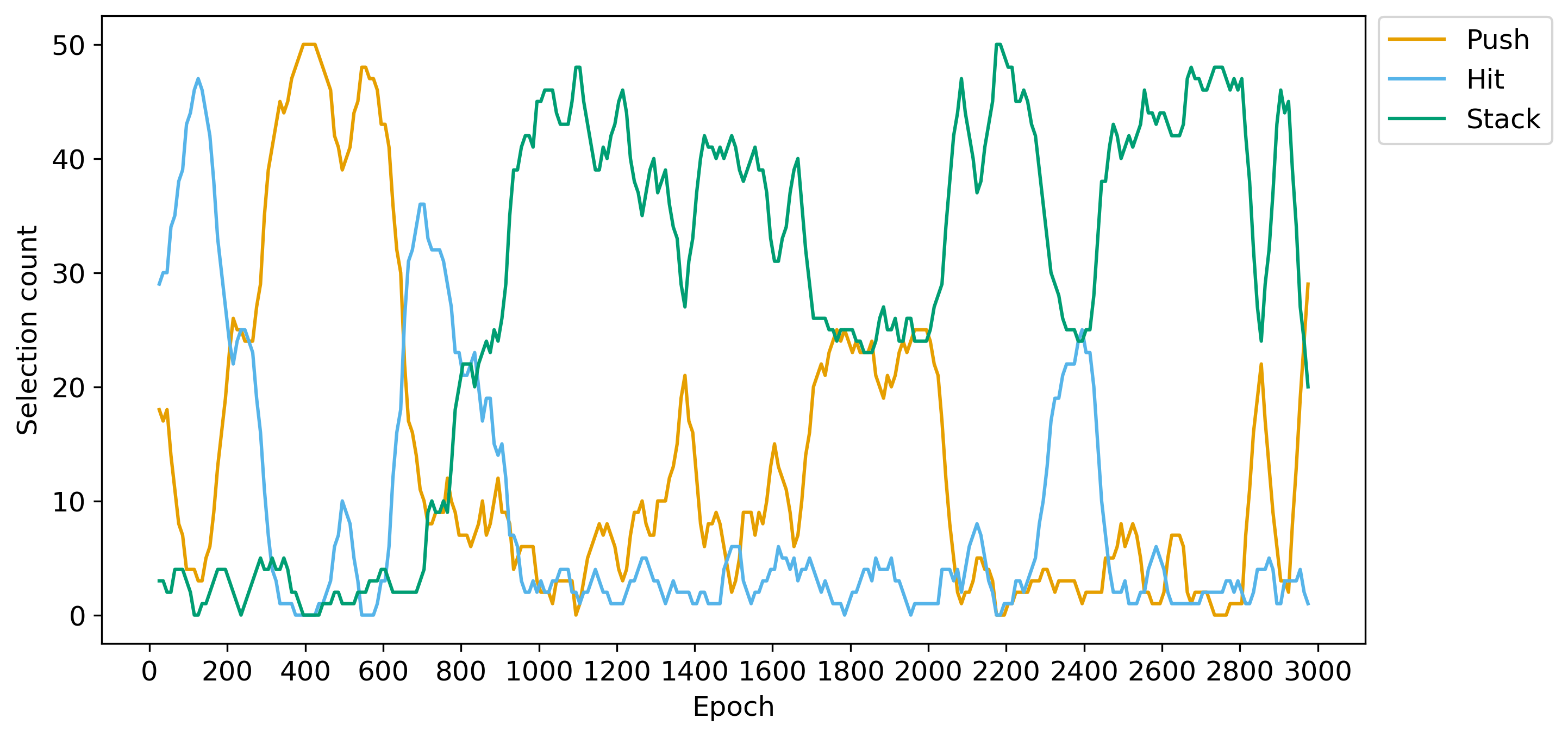}
    \label{fig:s2}}%
    \caption{Different task selection regimes by our IMTL-LP model during the training. Counts are calculated with a sliding window $w=50$ and a step size $s=10$.} 
    \label{fig:selections}
\end{figure*}
\bibliographystyle{elsarticle-num}
\bibliography{refs}

\end{document}